\documentclass{tlp}
\def\smallromani{\renewcommand{\theenumi}{\roman{enumi}}}

\usepackage{latexsym}
\usepackage{alltt}
\usepackage{epsfig}

\newcommand{\eclipse}{ECL$^i$PS$^e$}

\newcommand{\ra}{\mbox{$\:\rightarrow\:$}}

\newcommand{\La}{\mbox{$\:\Leftarrow\:$}}
\newcommand{\Ra}{\mbox{$\:\Rightarrow\:$}}

\newcommand{\A}{\mbox{$\ \wedge\ $}}

\newcommand{\sse}{\mbox{$\:\subseteq\:$}}

\newcommand{\po}{\mbox{$\ \sqsubseteq\ $}}

\newcommand{\LL}{\mbox{$\ldots$}}

\newcommand{\C}[1]{\mbox{$\{{#1}\}$}}           

\newcommand{\NI}{\noindent}
\newcommand{\HB}{\hfill{$\Box$}}

\newcommand{\III}{\vspace{3 mm}}
\newcommand{\II}{\vspace{2 mm}}

\newtheorem{theorem}{Theorem}[section]
\newtheorem{definition}[theorem]{Definition}
\newtheorem{lemma}[theorem]{Lemma}
\newtheorem{example}[theorem]{Example}
\newtheorem{note}[theorem]{Note}

\newcommand{\szkew}[1]{\relax \setbox0=\hbox{\kern -24pt
$\displaystyle#1$\kern 0pt }%
\box0}
{\catcode`\@=11 \global\let\ifjusthvtest@=\iffalse}

\newcounter{oldmycaption}

\newcommand{\p}[2]{\langle #1 \ ; \ #2 \rangle}

\title[Constraint Programming viewed as Rule-based Programming]
{Constraint Programming viewed as Rule-based Programming}

\author[Krzysztof R. Apt and Eric Monfroy]
{Krzysztof R. Apt\\
CWI\\
P.O. Box 94079, 1090 GB Amsterdam, the Netherlands\\
and University of Amsterdam, the Netherlands\\
\email{K.R.Apt@cwi.nl}
\and Eric Monfroy\\
Universit\'{e} de Nantes \\
Institut de Recherche en Informatique de Nantes (IRIN) \\
2, rue de la Houssinière \\
BP 92208, 44322 Nantes Cedex 03, France \\
\email{Eric.Monfroy@irin.univ-nantes.fr}
}

\begin{document}

\maketitle
\begin{abstract}

We study here a natural situation when constraint programming can be
entirely reduced to rule-based programming. 
To this end we explain first how one can compute on constraint satisfaction problems using 
rules represented by simple first-order formulas.
Then we consider
constraint satisfaction problems that are based on predefined,
explicitly given constraints.
To solve them we first derive rules from these explicitly given constraints
and limit the computation process to a repeated application
of these rules, combined with labeling.

We consider here two types of rules. The first type, that we call
equality rules, leads to a new notion of local consistency,
called {\em rule consistency} that turns out to be weaker than arc
consistency for constraints of arbitrary arity (called hyper-arc
consistency in \cite{MS98b}).  For Boolean constraints rule
consistency coincides with the closure under the well-known
propagation rules for Boolean constraints.  The second type of rules,
that we call membership rules, yields a rule-based characterization of
arc consistency.

To show feasibility of this rule-based approach to constraint
programming we show how both types of rules can be automatically
generated, as {\tt CHR} rules of \cite{fruhwirth-constraint-95}. This
yields an implementation of this approach to programming by means of
constraint logic programming.

We illustrate the usefulness of this approach to constraint
programming by discussing various examples, including Boolean
constraints, two typical examples of many valued logics, constraints
dealing with Waltz's language for describing polyhedral scenes, and
Allen's qualitative approach to temporal logic.
\II

\NI {\em Note.} A preliminary version of this article
appeared as \cite{AM99}. 
In this version we also present a framework for computing with rules
on constraint satisfaction problems and discuss in detail the results
of various experiments.
\end{abstract}
\section{Introduction}

\subsection{Background}

This paper is concerned with two styles of programming:
constraint programming and rule-based programming.

In constraint programming the programming process is limited to a
generation of constraints and a solution of the so obtained constraint
satisfaction problems (CSP's) by general or domain dependent methods.

In rule-based programming the programming process consists of a
repeated application of rules. A theoretical basis for this
programming paradigm consists of so-called production rules that were
introduced in the seventies, see, e.g., \cite{LS98}[pages 171-186],
though
the idea goes back to the works of A. Thue and of E. Post in first
half of twentieth century.  The production rules are condition-action
pairs, where the condition part is used to determine whether the rule
is applicable and the action part defines the action to be taken. The
most known programming language built around this programming paradigm
was OPS5 of \cite{For81}.

Recently, there has been a revival of interest in rule-based programming
in the context of constraint programming. The earliest example is
the {\tt CHR} language of \cite{fruhwirth-constraint-95} that is a part
of the \eclipse{} system.  (For a more recent and more complete
overview of {\tt CHR} see \cite{FruehwirthJLP98}.) The
{\tt CHR} rules extend the syntax of constraint logic programming by
allowing two atoms in the conclusion and employing guards. These rules
are predominantly used to write constraint solvers.

Another example of a programming language in which rules play an
important role is {\sf ELAN}. It offers a logical environment for
specifying and prototyping deduction systems by means of conditional
rewrite rules controlled by strategies.  {\sf ELAN} is used to support
the design of various rule-based algorithms such as constraints
solvers, decision procedures, theorem provers, and algorithms
expressed in logic programming languages, and to provide a modular
framework for studying their combinations. A general overview of {\sf
  ELAN} can be found in \cite{ELANoverview}, whereas
\cite{ELANconstraints} and \cite{Cas98} (to which we shall return in
Section \ref{sec:related}) describe applications of {\sf ELAN} to
constraint programming and constraint solving.

Also, in the hybrid functional and object-oriented language
programming language CLAIRE of \cite{CL96} rules are present.
CLAIRE was designed to apply constraint programming techniques for
operations research problems.  The rule-based programming is supported
by means of production rules that can be naturally used to
express constraint propagation.

It is useful to mention here that also logic programming and
constraint logic programming are rule-based formalisms.  However,
these formalisms use rules differently than rule-based programming
described above.  This distinction is usually captured by referring to
{\em forward chaining\/} and {\em backward chaining}.  In rule-based
programming, as discussed above, forward chaining is used while in
logic programming and constraint logic programming backward chaining
is employed. Intuitively, forward chaining aims at a simplification of
the considered problem and it maintains equivalence, while backward
chaining models reasoning by cases, where each case is implicitly
represented by a different rule.  Both forms of chaining can be
combined and in fact such a combination is realized in the {\tt CHR}
language, in which the {\tt CHR} rules model forward chaining while
the usual Prolog rules model backward chaining.

\subsection{Overview of Our Approach}

The traditional way of solving CSP's consists of combining constraint
propagation techniques with search.  Constraint propagation aims at
reducing a CSP to an equivalent one but simpler.  In case of finite
domains the most basic approach to search consists of labeling, a
repeated enumeration of the domains of the successive variables.

The aim of this paper is to show that constraint programming can be
entirely rendered by means of rule-based programming.  To this end we
provide a framework in which one computes on CSP's by means of rules
represented by simple first-order formulas. In this approach the
constraint propagation is achieved by repeated application of the
rules while search is limited to labeling.  This yields a framework
for constraint programming more related to logic than the usual one
based on algorithms achieving local consistency.

The rules we shall consider are implications built out of simple
atomic formulas. In our study we focus on two types of rules.
The first type,
that we call {\em equality rules}, are of the form
\[
x_1 = s_1, \LL, x_n = s_n \ra y \neq t
\]
where
$x_1, \LL, x_n,y$ are variables and $s_1, \LL, s_n, t$ are elements
of the respective variable domains. The computational interpretation of such
a rule is:
\begin{quote}
if for $i \in [1..n]$ the domain of the variable $x_i$ equals the
singleton
$\C{s_i}$, then remove the element $t$ from the domain of $y$.
\end{quote}

The second type of rules, that we call {\em membership rules}, are
of the form
\[
x_1 \in S_1, \LL, x_n \in S_n \ra y \neq t
\]
where

\begin{itemize}
\item 
$x_1, \LL, x_n$ are variables and $S_1, \LL, S_n$ are subsets
of the respective variable domains,

\item  $y$ is a variable and $t$ is an element of its domain.
\end{itemize}

The computational interpretation of such
a rule is:

\begin{quote}
if for $i \in [1..n]$ the domain of the variable $x_i$
is included in the set $S_i$, then remove the element $t$ from the
domain of $y$.
\end{quote}

To illustrate the use of these rules we study CSP's that are built out of
predefined, explicitly given finite constraints.  Such CSP's often
arise in practice. Examples include Boolean constraints, constraints
dealing with Waltz's language for describing polyhedral scenes,
Allen's temporal logic, and constraints in any multi-valued logic.

To solve such CSP's we explore the structure of these explicitly given
constraints first.  This information is expressed in terms of valid
equality and membership rules.  The computation process for a CSP
built out of these constraints is consists of two phases: a
generation of the rules from the explicitly given constraints and a
repeated application of these rules, combined with labeling.

To characterize the effect of the generated equality and membership
rules we use the notions of local consistency. These notions
approximate in a loose sense the notion of ``global consistency'',
see, e.g., \cite{Tsa93}.  We show that the first type of rules leads
to a local consistency notion that turns out to be weaker than arc
consistency for constraints of arbitrary arity. We call it {\em rule
  consistency}.

When the original domains are all unary or binary, rule consistency
coincides with arc consistency.  When additionally the predefined
constraints are the truth tables of the Boolean connectives, these
rules are similar to the well-known Boolean propagation rules, (see,
e.g., \cite{FruehwirthJLP98}[page 113]).
As a side effect, this shows that
the Boolean propagation rules characterize arc consistency.  Rule
consistency is thus a generalization of the Boolean propagation
to non-binary domains.

We also show that the membership rules lead to a notion of local
consistency that coincides with arc consistency. This yields
a rule-based implementation of arc consistency.

To show feasibility of this rule-based approach to constraint
programming, we automatically generate both types of rules, for an
explicitly given finite constraint, as rules in the {\tt CHR}
language. When combined with a labeling procedure such {\tt CHR}
programs constitute automatically derived decision procedures for the
considered CSP's, expressed on the constraint programming language
level.  In particular, we automatically generate the algorithms that
enforce rule consistency and arc consistency.

The availability of the algorithms that enforce rule consistency and
arc consistency on the constraint programming language level further
contributes to the automation of the programming process within the
constraint programming framework. In fact, in the case of such CSP's
built out of predefined, explicitly given finite constraints, the user
does not need to write one's own {\tt CHR} rules for the considered
constraints and can simply adopt all or some of the rules that are
automatically generated.  In the final example of the paper we also
show how using the equality rules and the membership rules, we can
implement more powerful notions of local consistency.

Alternatively, the generated equality rules and membership rules could be fed
into any of the generic {\em Chaotic Iteration\/} algorithms of
\cite{Apt99b} and made available in such systems as the ILOG solver.
This would yield rule consistency and an alternative implementation
of arc consistency.

The algorithms that for an explicitly given finite constraint generate
the appropriate rules that characterize rule consistency and arc
consistency have (unavoidably) a running time that is exponential in
the number of constraint variables and consequently are in general
impractical.

To test the usefulness of these algorithms for small finite domains we
implemented them in \eclipse{} and successfully used them on several
examples including the ones mentioned above. The fact that we could
handle these examples shows that this approach is of practical value
and in particular can be used to automatically derive practical
decision procedures for constraint satisfaction problems defined over
small finite domains.  Also it shows the usefulness of the {\tt CHR}
language for an automatic generation of constraint solvers and of
decision procedures.

\subsection{Organization of the Paper}

The rest of the paper is organized as follows. In the next section we
clarify the syntax of the rules and explain how one can compute with
them. In
Section \ref{sec:example} we illustrate the use
of these computations by means of an example.
In Section \ref{sec:outcomes} we prove that the outcomes
of the computations we are interested in are unique. 
In Section \ref{sec:semantics} we introduce some semantic aspects
of the rules and in Section \ref{sec:predefined} we formalize the concept
of a CSP is built out of predefined constraints.  Next, in Section
\ref{sec:rule-consistency} we introduce the notion of rule consistency
and discuss an algorithm that can be used to generate the minimal set
of rules that characterize this notion of local consistency. Then, in
Section \ref{sec:relation-arc-consistency} we compare rule consistency
with arc consistency. In Section \ref{sec:inclusion-rule-consistency} we
study  membership rules and
discuss an algorithm analogous to the one of
Section~\ref{sec:rule-consistency}. This entails a notion of local
consistency that turns out to be equivalent to arc consistency.

In Section \ref{sec:examples} we discuss the implementation
of both algorithms. They generate from an explicit representation of a
finite constraint a set of {\tt CHR} rules that characterize
respectively rule consistency and arc consistency. We also illustrate
the usefulness of these implementations by means of several examples.
Finally, in Section \ref{sec:related} we discuss other works in 
which  a link was made between constraint programming
and rule-based programming and in Section \ref{sec:conclusions}
we assess the merits of our approach. In the appendix we
summarize the tests carried out by means of our implementation
of both algorithms.  

\section{Computing with Rules}
\label{sec:computing}

In what follows we introduce specific type of rules and explain how
one can compute with them on constraint satisfaction problems.  First, 
we introduce constraints.

Consider a sequence of variables $X := x_1, \LL, x_n$
where $n \geq 0$, with respective domains $D_1, \LL, D_n$
associated with them.  So each variable $x_i$ ranges over the domain
$D_i$.  By a {\em constraint} $C$ on $X$ we mean a subset of $D_1
\times \LL \times D_n$.
Given an element $d := d_1, \LL, d_n$ of $D_1 \times \LL \times D_n$
and a subsequence $Y := x_{i_1}, \LL, x_{i_\ell}$ of $X$ we denote by
$d[Y]$ the sequence $d_{i_1}, \LL, d_{i_{\ell}}$. In particular, for a
variable $x_i$ from $X$, $d[x_i]$ denotes $d_i$.

Next, we define the rules we are interested in.

\begin{definition}

  \begin{itemize}
  \item Let $x$ be a variable, $a$ an element and $S$ a set.
By an {\em atomic formula\/} we mean one of the following
formulas: $x = a$, $x \neq a$, $x \in S$.

  \item By a {\em rule\/} we mean an expression of the form
$A_1, \LL, A_m \ra B_1, \LL, B_n$, where each $A_i$ and $B_j$
is an atomic formula.
\HB
  \end{itemize}
\end{definition}

In what follows a rule will be always associated with some constraint.
Then every atomic formula $x = a$ or $x \neq a$ (respectively, $x \in
S$) will be such that $a$ belongs to the domain of $x$ (respectively,
$S$ is a subset of the domain of $x$).

Subsequently, we explain how to compute using the rules in presence of
constraints.  First, we limit our considerations to the rules of the
form $A_1, \LL, A_m \ra x \neq a$.  We
need to explain how to turn the disequality formula into an action.
This is done by identifying the disequality $x \neq a$ with the
assignment $D_x := D_x -\C{a}$, where $D_x$ is the current domain of $x$. In
other words, we interpret $x \neq a$ as an action of removing the
value $a$ from the current domain of the variable $x$.

This leads us to the definition of an application of such a rule.
We need some semantic notions first.
\begin{definition}
Consider a constraint $C$ on a sequence of variables $X$,
a variable $x$ of $X$, and a tuple $d \in C$.

\begin{itemize}
\item
Given an atomic formula $A$ involving $x$ we
define the relation $\models_d A$ as follows:

  \begin{itemize}
  \item  $\models_d x = a$ iff $d[x] = a$,

  \item  $\models_d x \neq a$ iff $d[x] \neq a$,

  \item  $\models_d x \in S$ iff $d[x] \in S$.

  \end{itemize}

\item Given a sequence of atomic formulas ${\bf A} := A_1, \LL, A_m$
we define
$\models_d {\bf A}$ iff $\models_d A_i$ for all $i \in [1..m]$.
  \HB
\end{itemize}

\end{definition}

\begin{definition}
  Consider a constraint $C$ on a finite sequence of variables $X$
  and a rule of the form ${\bf A} \ra x_i \neq a$ involving only
  variables from $X$.

  Suppose that for all $d \in C$ we have $\models_d {\bf A}$.  Let
  $C'$ be the constraint obtained from $C$ by removing the element $a$
  from the domain $D_i$ of the variable $x_i$ and by removing from $C$
  all tuples $d$ such that $d[x_i] = a$.  Then we call the constraint
  $C'$ the {\em result of applying the rule} ${\bf A} \ra x_i \neq a$
  {\em to\/} $C$.  

  If $a \in D_i$, then we say that this is a {\em relevant
  application\/} of the rule ${\bf A} \ra x_i \neq a$ to $C$.  If $C'$
  coincides with $C$, we say that this application of the rule ${\bf
  A} \ra x_i \neq a$ to $C$ {\em maintains equivalence.}  
  \HB
\end{definition}

So the application of the rule ${\bf A} \ra x_i \neq a$ to a
constraint $C$ on the sequence $x_1, \LL, x_n$ of variables with
respective domains $D_1, \LL, D_n$ results in the constraint $C'$ on
the variables $x_1, \LL, x_n$ with respective domains $D_1, \LL,
D_{i-1}, D'_i, D_{i+1}, \LL, D_n$, where
\begin{itemize}
\item $D'_i = D_i - \C{a}$,

\item $C' = C \cap (D_1 \times \LL \times D_{i-1} \times D'_i \times
  D_{i+1}, \LL, D_n)$.
\end{itemize}

We say then that the constraint $C$ {\em is restricted to the domains}
$D_1, \LL, D_{i-1}, D'_i,$ $D_{i+1}, \LL, D_n$.

Now that we defined the result of a single application of a rule we
proceed to define computations. To this end we first
introduce constraint satisfaction problems.

By a {\em constraint satisfaction problem}, in short CSP, we mean a
finite sequence of variables $X$ with respective domains ${\cal
  D}$, together with a finite set $\cal C$ of constraints, each on a
subsequence of $X$. We write it as $\p{{\cal C}}{x_1 \in D_1,
  \LL, x_n \in D_n}$, where $X := x_1, \LL, x_n$ and ${\cal D} :=
D_1, \LL, D_n$.

By a {\em solution\/} to $\p{{\cal C}}{x_1 \in D_1, \LL, x_n \in D_n}$
we mean an element $d \in D_1 \times \LL \times D_n$ such that for
each constraint $C \in {\cal C}$ on a sequence of variables $X$ we
have $d[X] \in C$.  We call a CSP {\em consistent\/} if it has a
solution.  Two CSP's with the same sequence of variables are called
{\em equivalent\/} if they have the same set of solutions.

We now modify the definition of an application of a rule
to a constraint to an application of a rule to a CSP.
To this end we attach each rule to a constraint to which
it is supposed to be applied. Even though the constraints
change during the computations we consider, it will be always
clear from the context to which constraint a given rule is attached.

\begin{definition}
Consider a CSP ${\cal P}$ and a rule ${\bf A} \ra x_i \neq a$
attached to a constraint $C$ of ${\cal P}$. Suppose
that for all $d \in C$ we have $\models_d {\bf A}$.
Define a CSP ${\cal P}'$ on the same variables as ${\cal P}$ as follows:

\begin{itemize}

\item the domain of $x_i$ in ${\cal P}'$ equals $D_i - \C{a}$,
where $D_i$ is the domain of $x_i$ in ${\cal P}$,

\item the domains of other variables in ${\cal P}'$ are the
same as in ${\cal P}$,

\item the constraints of ${\cal P}'$ are obtained by restricting
the constraints of ${\cal P}$ to the new domains.

\end{itemize}

We say then that the CSP ${\cal P}'$ is the {\em result of applying
the rule} ${\bf A} \ra x_i \neq a$ {\em to\/} ${\cal P}$. If $a \in
D_i$, then we say that this is a {\em relevant application\/} of the
rule ${\bf A} \ra x_i \neq a$ to ${\cal P}$.
\HB
\end{definition}

Finally, we introduce the crucial notion of a computation.

\begin{definition}
  Consider a set of rules ${\cal R}$ of the form ${\bf A} \ra x \neq a$
  and an initial CSP ${\cal P}$.  By a {\em computation by means of
  ${\cal R}$ starting at ${\cal P}$} we mean a maximal sequence of
  CSP's ${\cal P}_1, \LL, {\cal P}_i, \LL$ such that each ${\cal
  P}_{j+1}$ is the result of a relevant application of a rule from
  ${\cal R}$ to ${\cal P}_{j}$.  \HB
\end{definition}

Note that when the set of rules ${\cal R}$ is finite or when all
domains in ${\cal P}$ are finite, all computations starting at ${\cal
  P}$ are finite. The reason is that in each of these two cases the
number of elements mentioned in the conclusions of the rules in ${\cal
  R}$ is finite.  But each element can be removed from a domain only
once and we insist that in computations
each rule application is relevant, from which
the claim follows. If a computation is finite, then no application of
a rule from the considered set of rules ${\cal R}$ to the final CSP is
relevant, i.e., this final CSP is {\em closed under the rules in ${\cal R}$}.  

The computations are a means to reduce the domains of the variables
while preserving the equivalence of the considered CSP.  The
computations here considered are in general insufficient for solving a
CSP and in the case of CSP's with finite domains they have to be
combined with labeling.  Labeling can be modeled in the above
rule-based framework by introducing a rule that splits a given CSP
into two, the union of which is equivalent to the given CSP. The
addition of such a rule to the considered framework leads to no
conceptual difficulties and is omitted. On the other hand, various
forms of labeling strategies, like the one in which variable with the
smallest domain is chosen first, cannot be captured on this level.

The above string of definitions allowed us to define computations in
which the actions are limited to the applications of rules of the form
${\bf A} \ra x \neq a$ acting on CSP's.

By limiting our attention to such type of rules we do not lose any
expressiveness. Indeed, consider first a rule of the form 
${\bf A} \ra x= a$.  To compute with it we interpret the equality $x = a$ 
as the assignment $D_x := D_x \cap \C{a}$, where $D_x$ is the current 
domain of $x$.
Then each rule ${\bf A} \ra x = a$ is equivalent to the conjunction of
the rules of the form ${\bf A} \ra x \neq b$ with $b \in D -\C{a}$,
where $D$ is the original domain of $x$.

Next, consider a rule of the form ${\bf A} \ra x \in S$.  To compute
with it we interpret the atomic formula $x \in S$ as the assignment
$D_x := D_x \cap S$, where $D_x$ is the current domain of $x$.  Then each rule
${\bf A} \ra x \in S$ is equivalent to the conjunction of the rules of
the form ${\bf A} \ra x \neq b$ with $b \in D - S$, where $D$ is the
original domain of $x$.

Finally, each rule of the form ${\bf A} \ra B_1, \LL, B_m$ is
equivalent to the conjunction of the rules of the form ${\bf A} \ra
B_i$ for $i \in [1..m]$.

Note that the rules of the form $x_1 = a_1, \LL, x_n = a_n \ra y \neq
b$ are more expressive than so-called {\em dependency rules\/} of
database systems (see, e.g., \cite{Ull88}). These are rules of the form
$x_1 = a_1, \LL, x_n = a_n \ra y = b$. We just explained how to model
them by means of rules of the form $x_1 = a_1, \LL, x_n =
a_n \ra y \neq b$.

However, modeling in the other direction is not possible, as can be
seen by taking the variables $x,y$, each with the domain $\C{0,1,2}$,
and the constraint $C$ on $x,y$ represented by the following table:

\[
\begin{array}{|l|l|}
\hline
x & y \\ \hline \hline
0 & 1 \\
0 & 0 \\
2 & 2 \\ \hline
\end{array}
\]
Then the rule $x = 0 \ra y \neq 2$ is not
equivalent
to a conjunction of the dependency rules.

\section{An Example}
\label{sec:example}

We now show how we can use the rules for computing by means of
an example kindly provided to us by Victor Marek.
We solve here a simple logic puzzle from \cite{Fle00}.

Below, given a set of variables $x_1, \LL, x_n$, each with the
domain $D$ we denote the following set of rules:
\[
\C{x_i = a \ra x_j \neq a \mid i, j \in [1..n], i \neq j, \ a \in D} 
\]
by $\mathit{all\_different}(x_1, \LL, x_n)$.  These rules formalize
the requirement that the variables $x_1, \LL, x_n$ are all different.
The puzzle is as follows.

\begin{quote}
{\em To stave off boredom on a rainy
Saturday afternoon, Ms. Rojas invented a game for Denise and her two
other children to play. Each child selected a different household object
(no two of which were in the same room) to describe to the others, who
would try to guess the item and its location in the house. Can you match
each child with the item he or she selected to describe, as well as the
room of the house (one is the living room) in which each is located?
}
\end{quote}

Here are  the clues provided:
\begin{enumerate}
\item  {\em The three children are Byron, the child who selected the
book, and the one whose item is in the den;}
\item {\em The rug is in the dining room;}
\item {\em Felicia selected the picture frame.}
\end{enumerate}

To solve this puzzle we use nine variables,
\begin{itemize}
\item $child_1, child_2, child_3$, to denote the three children, Byron, Denise and Felicia,
\item $room_B, room_D, room_F$, to denote the rooms of, respectively,
Byron, Denise and Felicia,
\item $item_B, item_D, item_F$, to denote the objects selected by,
  respectively, Byron, Denise and Felicia.
\end{itemize}

We postulate that the domain of $child_1$ is $\C{Byron}$, of $child_2$
is $\C{Denise}$ and of $child_3$ is $\C{Felicia}$.  Next, we assume
that each $room_i$ variable has the set $\C{\mathrm{den, dining, living}}$ as
its domain and each $item_i$ variable has the set $\C{\mathrm{book, frame,
  rug}}$ as its domain.

The initial set up of the story is formalized by the following rules:

\begin{itemize}
\item $\mathit{all\_different}(room_B, room_D, room_F)$,
\item $\mathit{all\_different}(item_B, item_D, item_F)$.
\end{itemize}

This yields 36 rules but we shall group the rules with the same
premise, so we shall actually have 18 rules. The rules we shall need below
will be, from the first set:
\medskip

(r1) $room_B = \mathrm{dining} \ra room_D \neq \mathrm{dining}, room_F \neq \mathrm{dining}$,

(r2) $room_B = \mathrm{living} \ra room_D \neq \mathrm{living}, room_F \neq \mathrm{living}$,

(r3) $room_D = \mathrm{living} \ra room_B \neq \mathrm{living}, room_F \neq \mathrm{living}$,
\medskip

\NI
and from the second set:
\medskip

(r4) $item_B = \mathrm{rug} \ra item_D \neq \mathrm{rug}, item_F \neq \mathrm{rug}$, 

(r5) $item_F = \mathrm{frame} \ra item_B \neq \mathrm{frame}, item_D \neq \mathrm{frame}$.
\medskip

Next, the first clue is formalized by means of eight rules out of which
the only ones of relevance below will be 
\medskip

(c1.1) $\ra item_B \neq \mathrm{book}$,

(c1.2) $item_D = \mathrm{book} \ra room_D \neq \mathrm{den}$.

\medskip
The second clue is formalized by means of six rules out of which the
only one of relevance below will be 
\medskip

(c2) $item_B = rug \ra room_B \neq \mathrm{den}, room_B \neq \mathrm{living}$.
\medskip

Finally, the third clue is formalized by means of two rules:
\medskip

(c3.1) $\ra item_F \neq \mathrm{rug}$,

(c4.1) $\ra item_F \neq \mathrm{book}$.
\medskip

So in total we have 34 rules but we shall use below only the ten rules
made explicit. The initial CSP has nine variables as introduced
above and one single ``universal'' constraint that consists of the
Cartesian product of all the variable domains.  The computation
consists of twelve steps and proceeds as follows.

\begin{enumerate}
\item  Using the rule (c3.1) the domain of $item_F$ is limited  to
$\{\mathrm{book}, \mathrm{frame}\}$.

\item  Using the rule (c3.2) the domain of $item_F$ is further limited to
$\{\mathrm{frame}\}$. Thus $item_F = \mathrm{frame}$ is established.

\item  Using the rule (r5) and the fact $item_F = \mathrm{frame}$
just established the domain of $item_B$ is limited to
$\{\mathrm{book}, \mathrm{rug}\}$.

\item  Using the rule (c1.1) the domain of $item_B$ is further limited to
$\{\mathrm{rug}\}$. Thus $item_B = \mathrm{rug}$ is established.

\item  Using the rule (r5) and the fact $item_F = \mathrm{frame}$
established in step 2 the domain of $item_D$ is limited to
$\{\mathrm{book}, \mathrm{rug}\}$.

\item Using the rule (r4) and the conclusion $item_B = \mathrm{rug}$
  of step 4 the domain of $item_D$ is further limited to
  $\{\mathrm{book}\}$. Thus $item_D = \mathrm{book}$ is established.

\item Using the rule (c2) and the fact $item_B = \mathrm{rug}$
  established in step 4 the domain of $room_B$ is limited to
  $\{\mathrm{dining},\mathrm{living}\}$.

\item Again using the rule (c2) and the fact $item_B = \mathrm{rug}$
  established in step 4 the domain of $room_B$ is further limited to
  $\{\mathrm{dining}\}$. Thus $room_B = \mathrm{dining}$ is
  established.
  
\item Using the fact $room_B = \mathrm{dining}$ established in step 8
  and the rule (r1) the domain of $room_D$ is limited to
  $\{\mathrm{den},\mathrm{living}\}$.

\item Using the fact $item_D = \mathrm{book}$ established in step 6 and 
the rule (c1.2), the domain of $room_D$ is further limited to
$\{\mathrm{living}\}$. Thus $room_D = \mathrm{living}$ is 
established.

\item Using the fact $room_B = \mathrm{dining}$ established in step 9
and the rule (r2) the domain of $room_F$ is limited to
$\{\mathrm{den},\mathrm{dining}\}$.

\item Using the fact $room_D = \mathrm{living}$ established in step 10 
and the rule (r3) the domain of $room_F$ is further limited to
$\{\mathrm{den}\}$. Thus $room_F = \mathrm{den}$ is established.
\end{enumerate}
At this stage one can check that the resulting CSP with all
singleton domains is closed under all 34 rules.
This yields the solution to the puzzle represented by the following table:

\begin{center}
\[
\begin{array}{|l|l|l|}
\hline
\mathrm{child} & \mathrm{room} & \mathrm{item} \\
\hline
\mathrm{Byron} & \mathrm{dining} &  \mathrm{rug} \\
\mathrm{Denise}   & \mathrm{living} & \mathrm{book} \\
\mathrm{Felicia}  & \mathrm{den} & \mathrm{frame} \\
\hline
\end{array}
\]
\end{center}

\section{Outcomes of Computations}
\label{sec:outcomes}

A natural question arises whether the outcome of computations using a
finite set of rules is unique.  The answer is positive.  To prove it
we need a lemma concerning iterations of inflationary and monotonic
functions.

\begin{definition}
  Consider a partial ordering $(D, \po )$ with the least element $\bot$
  and a finite set of functions $F := \C{f_1, \LL , f_k}$ on $D$.
 \begin{itemize}

\item 
By an {\em iteration of $F$\/} 
we mean an infinite sequence of values 
$d_0, d_1, \LL  $ defined inductively by
\[
d_0 := \bot,
\]
\[
d_{j} := f_{i_{j}}(d_{j-1}),
\]
where each $i_j$ is an element of $[1..k]$.

\item We say that an increasing sequence
$d_0 \: \po \: d_1 \: \po \: d_2 \: \LL$ of elements from $D$
{\em eventually stabilizes at d\/} if for some $j \geq 0$ we have
$d_i = d$ for $i \geq j$.

\item A function $f$ on $D$ is called {\em inflationary\/}
if $x \po f(x)$ for all $x$.

\item A function $f$ on $D$ is called {\em monotonic\/}
  if $x \po y$ implies $f(x) \po f(y)$ for  all $x, y$.  
\HB
\end{itemize}
\end{definition}

We now need the following lemma from \cite{Apt99c}.

\begin{lemma} \label{lem:stabilization}
  Consider a partial ordering $(D, \po )$ with the least element
  $\bot$ and a finite set of monotonic functions $F$ on $D$.
Suppose that an iteration of $F$ eventually stabilizes at a common
  fixpoint $d$ of the functions from $F$.  Then $d$ is the least
  common fixed point of the functions from $F$.  
\end{lemma}

It follows that all iterations of $F$ that eventually
stabilize at a common fixpoint stabilize at the same element.
We now prove the desired result.
\begin{theorem} \label{thm:confluence} 
  Fix an initial CSP ${\cal P}$.  Consider a finite set ${\cal R}$ of
  rules of the form ${\bf A} \ra x \neq a$.  Then all computations by
  means of ${\cal R}$ starting at ${\cal P}$ yield the same CSP.
\end{theorem}

\begin{proof}
We already noted in Section \ref{sec:computing}
that all such computations are finite.
Suppose that ${\cal P} := \p{{\cal C}}{x_1 \in D_1, \LL, x_n \in D_n}$.
We consider now the following partial ordering $(D, \po )$.  
The elements of $D$ are the sequences
$(E_1, \LL, E_n)$ such that $E_i \sse D_{i}$
for $i \in [1..n]$, ordered componentwise w.r.t. the reversed subset
ordering $\supseteq$.  So $(D_1, \LL, D_n)$ is the least element $\bot$
in this ordering and 

\[
\mbox{$(E_1, \LL, E_n) \po (F_1, \LL, F_n)$ iff
$E_i \supseteq F_i$ for $i \in [1..n]$.}
\]

We replace in each rule each premise atom $x_i = a$ by $x_i \in \C{a}$
and $x_i \neq a$ by $x_i \in D_i - \C{a}$.  Since for all $d \in D_1
\times \LL \times D_n$ we have $\models_d x_i = a$ iff $\models_d x_i
\in \C{a}$ and $\models_d x_i \neq a$ iff $\models_d x_i \in D_i -
\C{a}$, it follows that the applications of the original and of the
resulting rules coincide. This allows us to confine our attention to
the rule each premise of which is of the form $z \in S$.

Consider now a membership rule $z_1 \in S_1, \LL, z_m \in S_m \ra y
\neq a$ associated with a constraint $C$ from ${\cal C}$ defined on a
set of variables $Y$. We interpret this rule as a function on the just
defined set D as follows.

First, denote by $\bar{C}$ the extension ``by padding'' of $C$ to all
the variables $x_1, \LL, x_n$, i.e. $\bar{C} \sse D_1 \times \LL
\times D_n$ and $d \in \bar{C}$ iff $d[Y] \in C$.  Next, given a
constraint $E$ and its variable $z$ denote the set $\C{d[z] \mid d \in
  E}$ by $E[z]$.  Finally, assume for simplicity that $y$ is $x_n$.

The function $f$ that corresponds to the rule $z_1 \in S_1,
\LL, z_m \in S_m \ra y \neq t$ is defined as follows:
\[
f(E_1, \LL, E_n) := \left \{ \begin{array}{ll}
 (E_1, \LL, E_n - \C{a}) & \mbox{if $(\bar{C} \cap (E_1 \times \LL \times E_n))[z_i] \sse S_i$} \\
                         & \mbox{for $i \in [1..m]$}, \\
 (E_1, \LL, E_n)                            & \mbox{otherwise. }
                                      \end{array}
                            \right.
\]

Denote the set of so defined functions by $F$.
By definition each function $f \in F$ is inflationary and monotonic
w.r.t. the componentwise reversed subset ordering $\supseteq$. 

Now, there is a one-one correspondence between the common fixpoints of
the functions from $F$ at which the iterations of $F$ eventually
stabilize and the outcomes of the computations by means of ${\cal R}$
starting at ${\cal P}$. In this correspondence a common fixpoint
$(E_1, \LL, E_n)$ is related to the CSP
$\p{{\cal C}'}{x_1 \in E_1, \LL, x_n \in E_n}$
closed under the rules of ${\cal R}$, where 
${\cal C}'$ are the constraints from ${\cal C}$ restricted to the domains
$E_1, \LL, E_n$. The conclusion now follows by Lemma
\ref{lem:stabilization}.
\end{proof}

\section{Semantic Aspects of Rules}
\label{sec:semantics}

We now introduce a number of semantic notions concerning rules.

\begin{definition}
\label{def:semantic}
%
Consider a constraint $C$.
\begin{itemize}
\item We say that the rule
${\bf A} \ra {\bf B}$ is {\em valid for $C$\/}
if for all tuples $d \in C$
\[
\mbox{$\models_d {\bf A}$  implies $\models_d {\bf B}$.}
\]

\item We say that the constraint $C$ is {\em closed under the rule
${\bf A} \ra {\bf B}$} if
\[
\mbox{($\models_d {\bf A}$ for all tuples  $d \in C$) implies
($\models_d {\bf B}$ for all tuples  $d \in C$}).
\]

\item We say that the rule ${\bf A} \ra {\bf B}$ is {\em feasible for\/} $C$ 
if for some tuple $d \in C$ we have $\models_d  {\bf A}$.

\item We say that the rule ${\bf A} \ra {\bf B}$ for
the constraint $C$ {\em extends\/} the rule
${\bf A}' \ra {\bf B}$ if ${\bf A}$ contains all variables of ${\bf A'}$ 
and for all tuples $d \in C$
\[
\mbox{$\models_d {\bf A}$  implies $\models_d {\bf A}'$.}
\]

\item Given a set of rules ${\cal R}$, we call a rule {\em minimal in
${\cal R}$\/} if it is feasible and it does not properly extend a
valid rule in ${\cal R}$.
\HB
\end{itemize}

\end{definition}

To illustrate them consider the following example.

\begin{example} \label{exa:conjunction}
  Take as a constraint the ternary relation that represents the
  conjunction $and(x, y, z)$. It can be viewed as the
  following table:

\[
\begin{array}{|l|l|l|}
\hline
x & y & z \\ \hline \hline
0 & 0 & 0 \\
0 & 1 & 0 \\
1 & 0 & 0 \\
1 & 1 & 1 \\ \hline
\end{array}
\]
In other words, we assume that each of the variables $x,y,z$
has the domain $\C{0,1}$ and view
$and(x,y,z)$ as the constraint on $x,y,z$ that consists of
the above four triples.

It is easy to see that the rule $x = 0 \ra z \neq 1$ is valid
for $and(x,y,z)$. Further, the rule $x = 0, y = 1 \ra z \neq 1$
extends the rule $x = 0 \ra z \neq 1$ and is also valid for
$and(x,y,z)$. However, out of these two rules only $x = 0 \ra
z \neq 1$ is minimal.

Finally, both rules are feasible, while the rules $x = 0, z = 1 \ra y
\neq 0$ and $x = 0, z = 1 \ra y \neq 1$ are not feasible.  
\HB
\end{example}

Note that the definition of an application of a rule is so
designed that a link with semantics is kept in the following sense: if 
a rule $r$ is valid for a constraint $C$, then $C$ is closed under $r$.
Rules that are not feasible are trivially valid.
Note also that a rule that extends a valid rule is valid, as well.  So
validity extends ``upwards''.

Note the use of the condition
``${\bf A}$ contains all variables of ${\bf A'}$''
in the definition of the relation ``the rule ${\bf A} \ra {\bf B}$
extends the rule ${\bf A}' \ra {\bf B}$''.  Without it we would have
the following paradoxical situation.  Consider the variables $x,y,z$,
all on the domains $\C{0,1}$, and the constraint $C$ on $x,y,z$ 
defined by $C := \C{(0,0,0), (1,0,1)}$.
Then the rules $x = 0 \ra z \neq 1$ and $y = 0 \ra z \neq 1$ are
both valid for $C$ and for all $d \in C$ we have  that $\models_d x = 0$
implies  $\models_d y = 0$. So without the mentioned condition
we would have that the rule $x = 0 \ra z \neq 1$ extends the
rule $y = 0 \ra z \neq 1$, which would imply that the first rule
is not minimal.

In the sequel the following observation will be useful.

\begin{note} \label{not:rule-cons}
Consider two finite and non-empty constraints $C$ and $E$ such that $C
\sse E$ and a set of rules ${\cal R}$.  Then $C$ is closed under all
valid rules from ${\cal R}$ for $E$ iff it is closed under all minimal
valid rules in ${\cal R}$ for $E$.
\end{note}

\begin{proof}
Suppose that $C$ is closed under all minimal
valid rules in ${\cal R}$ for $E$.
Take a rule $r$ from  ${\cal R}$ that is valid for $E$. 
\II

\NI
{\em Case 1}. $r$ is feasible for $E$.

Then, because $E$ is finite, $r$ extends some minimal valid rule $r'$
in ${\cal R}$ for $E$.  But $C$ is closed under $r'$, so it is closed
under $r$, as well.  
\II

\NI
{\em Case 2}. $r$ is not feasible for $E$.

Then $r$ is not feasible for $C$ either since $C \sse E$.
Consequently, since $C$ is non-empty, $C$ is closed under $r$.
\end{proof}

In what follows we confine our attention to computations
involving two types of rules:

\begin{itemize}
\item {\em equality rules\/}:
these are rules of the form $x_1=s_1, \LL, x_m=s_m \ra y \neq a$;
we abbreviate them to $X = s \ra y \neq a$,
where $X = x_1, \LL, x_m$ and $s = s_1, \LL s_m$,
\item {\em membership rules\/}:
these are rules of the form $x_1 \in S_1, \LL, x_m \in S_m \ra y \neq a$;
we abbreviate them to $X \in S \ra y \neq a$,
where $X = x_1, \LL, x_m$ and $S = S_1, \LL S_m$.
\end{itemize}

By specializing in the last clause of Definition \ref{def:semantic},
defining a minimal rule, the set ${\cal R}$ of rules to the set of
equality rules and to the set of membership rules we obtain the
notions of a \emph{minimal equality rule} and of a \emph{minimal
  membership rule}.  For equality and membership rules the following
straightforward characterization of the ``extends'' relation will be
of use.

\begin{note} \label{not:rule-extends}
  \begin{enumerate}\smallromani
  \item An equality rule  $x_1=s_1, \LL, x_m=s_m \ra y \neq a$ extends an equality rule
 $z_1=t_1, \LL, z_n=t_n \ra y \neq a$ iff $z_1=t_1, \LL, z_n=t_n$ is a subsequence of
$x_1=s_1, \LL, x_m=s_m$.

\item A membership rule $x_1 \in S_1, \LL, x_m \in S_m \ra y \neq a$
  extends a membership rule $z_1 \in T_1, \LL, z_n \in T_n \ra y \neq
  a$ iff $z_1, \LL, z_n$ is a subsequence of $x_1, \LL, x_m$ 
and for each $i \in [1..n]$ we have $S_{\pi(i)} \sse T_i$, where $z_i$
  equals $x_{\pi(i)}$.
\HB
  \end{enumerate}

\end{note}

Given a CSP with finite domains we would like to solve it by
considering computations starting at it.  But
where do we get the rules from? Note that given a constraint $C$ and a
rule $r$ that is valid for $C$, the constraint $C$ is trivially closed
under $r$.  Consequently, an application of $r$ to $C$ is not
relevant, i.e., it does not affect $C$. So to obtain some change
we need to use rules that are not valid for the initial constraints.
This brings us to the notion of a CSP based on another one.

\section{CSP's Built out of Predefined Constraints}
\label{sec:predefined}

In the introduction we informally referred to the notion of a CSP
``being built out of predefined, explicitly given constraints.''  Let
us make now this concept formal.  We need two auxiliary notions first,
where in preparation for the next definition we already consider
constraints together with the domains over which they are defined.

\begin{definition}

  \begin{itemize}
  \item Given a constraint $C \sse D_1 \times \LL \times D_n$
and a permutation $\pi$ of $[1..n]$ we denote by
$C^{\pi}$ the relation defined by
\[
(a_1, \LL, a_n) \in C^{\pi} \mbox{ iff } (a_{\pi(1)}, \LL, a_{\pi(n)})
\in C
\]
and call it {\em a permutation of $C$}.
\item Given two constraints
$C \sse D_1 \times \LL \times D_n$ and $E \sse D'_1 \times \LL \times
D'_n$
we say that {\em $C$ is based on $E$\/} if
\begin{itemize}
\item $D_i \sse D'_i$ for $i \in [1..n]$,

\item $C = E \cap (D_1 \times \LL \times D_n)$.
\HB
\end{itemize}
  \end{itemize}
\end{definition}
So the notion of ``being based on'' involves
the domains of both constraints.
If $C$ is based on $E$, then $C$ is the restriction of
$E$ to the domains over which $C$ is defined.

\begin{definition} \label{def:based}

We assume that the ``predefined constraints'' are presented as a given
in advance CSP ${\cal BASE}$. Suppose that each constraint $C$ of 
a CSP ${\cal P}$ is based on a permutation of a constraint of  ${\cal BASE}$.
%
%
%
%
%
We say then that ${\cal P}$ is {\em based on\/}
${\cal BASE}$.
\HB
\end{definition}

In the above definition the use of permutations of constraints allows us
to abstract from the variable ordering used in ${\cal BASE}$.
The following example illustrates this notion.

\begin{example} \label{exa:full-adder}
Consider the well-known full adder circuit.
It is defined by the following formula:
\[
\begin{array}{l}
add(i_1,i_2,i_3,o_1,o_2) \equiv \\
~~        xor(i_1,i_2,x_1),
        and(i_1,i_2,a_1),
        xor(x_1,i_3,o_2),
        and(i_3,x_1,a_2),
        or(a_1,a_2,o_1),
\end{array}
\]
where $and, xor$ and $or$ are defined in the expected way.
We can view the original constraints as the following CSP:
\[
{\cal BOOL} := \p{and(x,y,z),  xor(x,y,z),  or(x,y,z)}
                 {x \in \C{0,1}, y \in \C{0,1}, z \in \C{0,1}}.
\]

${\cal BOOL}$ should be viewed just as an ``inventory'' of the
predefined constraints and not as a CSP to be solved.  Now, any query
concerning the full adder can be viewed as a CSP based on ${\cal
BOOL}$.  For example, in Section \ref{sec:examples} we shall consider
the query $add(1,x,y,z,0)$. It corresponds to the following CSP based
on ${\cal BOOL}$:
\begin{tabbing}
\ $\langle$ \= $xor(i_1,i_2,x_1), \ and(i_1,i_2,a_1), \
xor(x_1,i_3,o_2), \ and(i_3,x_1,a_2), \ or(a_1,a_2,o_1)\ ;$ \\
\> $i_1 \in \C{1},
                        i_2 \in \C{0,1},
                        i_3 \in \C{0,1},
                        o_1 \in \C{0,1},
                        o_2 \in \C{0},
                        a_1 \in \C{0,1},
                        a_2 \in \C{0,1},$ \\
\>  $x_1 \in \C{0,1} ~ \rangle$. 
\end{tabbing}
\HB
\end{example}

In what follows we consider computations that start with a CSP 
based on some CSP ${\cal BASE}$.  In these computations we wish
to maintain equivalence between the successive CSP's.
To this end the following simple observation is crucial.

\begin{note} \label{not:equivalence}
Consider two constraints $C$ and $E$ such that $C$ is based on 
$E$. Let ${\bf A} \ra x \neq a$ be a rule valid for $E$.
Then the application of  ${\bf A} \ra x \neq a$ to $C$
maintains equivalence. 
\end{note}
\begin{proof}
Assume that the rule ${\bf A} \ra x \neq a$ can be applied to $C$,
i.e., that for all $d \in C$ we have $\models_d  {\bf A}$.
Suppose now that the rule ${\bf A} \ra x \neq a$ to $C$ does not 
maintain equivalence. Then for some $d \in C$ we have $d[x] = a$.
$C$ is based on $E$, so  $d \in E$. By the validity of the rule
for $E$ we get $d[x] \neq a$. This yields a contradiction.
\end{proof}

This observation provides us with a way of maintaining equivalence
during a computation: it suffices to use at each step a rule that is
valid for a permutation $C^{\pi}$ of a constraint of ${\cal
BASE}$. Such a rule is then attached (i.e., applied) to the constraint
based on $C^{\pi}$.  This is what we shall do in the sequel.
Depending on the type of rules used we obtain in this way different notions of
local consistency.

\section{Rule Consistency}
\label{sec:rule-consistency}

In this section we consider a CSP ${\cal P}$ based on some finite CSP
${\cal BASE}$ and study computations that use exclusively equality
rules.  The rules are obtained from the constraints of ${\cal BASE}$;
each of them is valid for a permutation $C^{\pi}$ of a constraint $C$
of ${\cal BASE}$ and is applied to the constraint of ${\cal P}$ 
based on $C^{\pi}$.
By Note \ref{not:equivalence} the successive CSP's are all equivalent
to the initial CSP ${\cal P}$.  The computation ends when a CSP is
obtained that is closed under the rules used. This brings us to a
natural notion of local consistency expressed in terms of equality
rules.

\begin{definition} \label{def:rule-cons}
Consider a CSP ${\cal P}$ based on a CSP ${\cal BASE}$.  Let
$C$ be a constraint of ${\cal P}$.
For some constraint
$f(C)$ of ${\cal BASE}$ and a permutation $\pi$,
$C$ is based on $f(C)^{\pi}$.

\begin{itemize}

\item We call the constraint $C$ {\em rule consistent\/} (w.r.t.
${\cal BASE}$) if it is closed under all equality rules that are valid
for $f(C)^{\pi}$.

\item We call the CSP ${\cal P}$ {\em rule consistent\/} (w.r.t.  ${\cal BASE}$)
if all its constraints are rule consistent.
\HB
\end{itemize}
\end{definition}

In what follows we drop the reference to ${\cal BASE}$ if it is clear
from
the context.

\begin{example}
Take as the base CSP
\[
{\cal BASE} := \p{and(x,y,z)}{x \in \C{0,1}, y \in \C{0,1}, z \in
\C{0,1}}
\]
and consider the following four CSP's based on it:
\begin{enumerate}

\item $\p{and(x,y,z)}{x \in \C{0}, y \in D_y, z \in \C{0}}$,

\item $\p{and(x,y,z)}{x \in \C{1}, y \in D_y, z \in \C{0,1}}$,

\item $\p{and(x,y,z)}{x \in \C{0,1}, y \in D_y, z \in \C{1}}$,

\item $\p{and(x,y,z)}{x \in \C{0}, y \in D_y, z \in \C{0,1}}$,

\end{enumerate}
where $D_y$ is a subset of $\C{0,1}$.  We noted in Example
\ref{exa:conjunction} that the equality rule $x = 0 \ra z \neq 1$ is
valid for $and(x,y,z)$.  In the first three CSP's its only constraint
is closed under this rule, while in the fourth one it is not closed
since 1 is present in the domain of $z$ whereas the domain of $x$
equals $\C{0}$.  So the fourth CSP is not rule consistent.  One can
show that the first two CSP's are rule consistent, while the third one
is not rule consistent since it is not closed under the valid equality
rule $z = 1 \ra x \neq 0$.
\end{example}

When trying to generate all valid equality rules
Note \ref{not:rule-cons}
allows us to confine our attention to the minimal valid equality rules.
We now introduce an algorithm that given a finite constraint generates the
set of all minimal valid equality rules for it.  
We collect the generated rules in a list.  We denote below the empty
list by {\bf empty} and the result of insertion of an element $r$ into
a list $L$ by ${\bf insert}(r, L)$.

By an {\em assignment\/} to a sequence of variables $X$ we mean here
an element $s$ from the Cartesian product of the domains of variables
of $X$ such that for some $d \in C$ we 
have $d[X] = s$.  Intuitively, if we represent the constraint $C$ as a
table with rows corresponding to the elements (tuples) of $C$ and the
columns corresponding to the variables of $C$, then an assignment to
$X$ is a tuple of elements that appears in some row in the columns
that correspond to the variables of $X$.  This algorithm has the
following form, where we assume that the considered constraint $C$ is
defined on a sequence of variables {\em VAR\/} of cardinality $n$.
\II

\NI
{\sc Equality Rules Generation} algorithm

\begin{alltt}
\emph{L := {\bf empty}};
FOR \emph{i:= 0} TO \emph{n-1} DO
   FOR each subset \emph{X} of \emph{VAR} of cardinality \emph{i} DO
      FOR each assignment \emph{s} to \emph{X} DO
         FOR each \emph{y} in \emph{VAR-X} DO
            FOR each element \emph{d} from the domain of \emph{y} DO
               \emph{r :=  X = s} \ra \emph{y} \(\neq\) \emph{d};
               IF \emph{r} is valid for \emph{C}
                  and it does not extend an element of \emph{L}
                  THEN \emph{{\bf insert}(r, L)}
               END
            END
         END
      END
   END
END
\end{alltt}

The test that one equality rule does not extend another can be easily
implemented by means of Note \ref{not:rule-extends}.i.

The following result establishes correctness of this algorithm.

\begin{theorem} \label{thm:rga}
Given a constraint $C$ the {\sc Equality Rules Generation} algorithm produces
in {\it L} the set of all minimal valid equality rules for $C$.
\end{theorem}
\begin{proof}
  First note that in the algorithm all possible feasible equality
  rules are considered and in the list {\it L} only the valid equality
  rules are retained.  Additionally, a valid equality rule is retained
  only if it does not extend a rule already present in {\it L}.  

  Finally, the equality rules are considered in the order according to
  which those that use less variables are considered first. By virtue
  of Note \ref{not:rule-extends}.i this implies that if a rule $r_2$
  extends a rule $r_1$, then $r_1$ is considered first.  As a
  consequence precisely all minimal valid equality rules are retained
  in {\it L}.
\end{proof}

The above algorithm is completely straightforward and consequently
inefficient.  It is easy to see that given a constraint defined over
$n$ variables, $O(n \cdot 2^{n} \cdot d^{2})$ rules are considered in
it, where $d$ is the size of the largest variable domain.  This shows
that in practice this algorithm is impractical for large domains and
for constraints with many variables.  By representing the rules
explicitly one could improve the running time of this algorithm,
trading time for space.  Then the test that one rule does not extend
another could be eliminated from the algorithm by representing
explicitly the partial ordering defined by the relation ``equality
rule $r_1$ extends equality rule $r_2$''. Each time an equality rule
that is valid for $C$ would be found, all the rules that extend it
would be now disregarded in further considerations.  This would reduce
the number of rules considered and improve the average running time.
However, it is difficult to quantify the gain obtained and in the
worst case still all the rules would have to be considered.

In Section \ref{sec:examples} and the appendix we present some
empirical results showing when the {\sc Equality Rules Generation}
becomes infeasible.  

\section{Relating Rule Consistency to Arc Consistency}
\label{sec:relation-arc-consistency}

To clarify the status of rule consistency we compare it now
to the notion of arc consistency. This notion was introduced in
\cite{mackworth-consistency} for binary relations and was extended to
arbitrary relations in  \cite{MM88}. Let us recall the
definition.

\begin{definition} 
  \begin{itemize}
  \item We call a constraint $C$ on a sequence of variables $X$ {\em arc
      consistent\/} if for every variable $x$ in $X$ and an
    element $a$ in its domain there exists $d \in C$ such that $a =
    d[x]$.  That is, each element in each
    domain participates in a solution to $C$.

\item We call a CSP {\em arc consistent\/} if all its
constraints are arc consistent.
\HB
  \end{itemize}
\end{definition}

The following result relates  for constraints of arbitrary
arity arc consistency to rule consistency.

\begin{theorem} \label{thm:rulecons}
Consider a CSP $\cal P$ based on a CSP ${\cal BASE}$.
If $\cal P$ is arc consistent, then it is rule consistent
w.r.t. ${\cal BASE}$.
\end{theorem}
\begin{proof}
Assume that $\cal P$ is arc consistent.  Choose a constraint $C$ of
$\cal P$ and consider an equality rule $X = s \ra y \neq a$ that is valid for
$f(C)^{\pi}$, where $f$ and $\pi$ are as in Definition
\ref{def:rule-cons}.

Suppose by contradiction that $C$ is not closed
under this rule. So for $X := x_{1}, \LL, x_{k}$ and $s := s_1, \LL,
s_k$ the domain of each variable $x_{j}$ in $\cal P$ equals $\C{s_j}$
and moreover $a \in D$, where $D$ is the domain of the variable $y$ in
$\cal P$.

By the arc consistency of $\cal P$ there exists $d \in C$ such that
$d[y] = a$. Because of the form of the domains of the variables in
$X$, also $d[X] = s$ holds.  Additionally, because $\cal P$ is based
on ${\cal BASE}$, we have $d \in f(C)^{\pi}$.  But by assumption
the equality rule $X = s \ra y \neq a$ is valid for $f(C)^{\pi}$, so $d[y] \neq
a$.  A contradiction.
\end{proof}

The converse implication does not hold in general as the following
example shows.

\begin{example} \label{exa:not-arc-consistent}
Take as the base the following CSP
\[
{\cal BASE} := \p{C}{x \in \C{0,1,2}, y \in \C{0,1,2}}
\]
where the constraint $C$ on $x,y$ that equals the set $\C{(0,1), (1,0),
(2,2)}$.
So $C$ can be viewed as the following table:
\III

\[
\begin{array}{|l|l|} \hline
x & y \\ \hline \hline
0 & 1 \\
1 & 0 \\
2 & 2 \\ \hline
\end{array}
\]

Next, take for $D_1$ the set $\C{0,1}$ and $D_2$ the set $\C{0,1,2}$.
Then the CSP $\p{C \cap (D_1 \times D_2)}{x \in D_1, y \in D_2}$, so
$\p{\C{(0,1), (1,0)}}{x \in \C{0,1}, y \in \C{0,1,2}}$ is based on
${\cal BASE}$ but is not arc consistent since the value 2 in the
domain of $y$ does not participate in any solution. Yet, it is easy to
show that the only constraint of this CSP is closed under all equality 
rules that are valid for $C$.
\HB
\end{example}

We now show that if each domain has at most two elements, then the
notions of arc consistency and rule consistency coincide.  More
precisely, the following result holds.

\begin{theorem} \label{thm:rule-consistency}
  Let ${\cal BASE}$ be a CSP each domain of which is unary or binary.
  Consider a CSP $\cal P$ based on ${\cal BASE}$.  Then $\cal P$ is
  arc consistent iff it is rule consistent w.r.t. ${\cal BASE}$.
\end{theorem}
\begin{proof}
The ($\Ra$) implication is the contents of Theorem \ref{thm:rulecons}.

To prove the reverse implication suppose that some constraint $C$
of $\cal P$ is not arc consistent. We prove that then $C$ is not rule
consistent.

The constraint $C$ is on some variables $x_1, \LL, x_n$ with
respective domains $D_1, \LL, D_n$.  For some $i \in [1..n]$ some $a
\in D_i$ does not participate in any solution to $C$.

Let $D_{i_1}, \LL, D_{i_\ell}$ be the sequence of all domains in $D_1,
\LL, D_{i-1}, D_{i+1}, \LL, D_n$ that are singletons. Suppose that
$D_{i_j} := \C{s_{i_j}}$ for $j \in [1.. \ell]$ and let $X := x_{i_1},
\LL, x_{i_\ell}$ and $s := s_{i_1}, \LL, s_{i_\ell}$.

Consider now the equality rule $X = s \ra x_i \neq a$ and take $f(C)^{\pi}$,
where $f$ and $\pi$ are as in Definition \ref{def:rule-cons}.  For
appropriate domains $D'_1, \LL, D'_n$ of ${\cal BASE}$ we have
$f(C)^{\pi} \sse D'_1 \times \LL \times D'_n$.

Next, take some $d \in f(C)^{\pi}$ such that $d[X] = s$.  We show that
$d \in C$. Since $C = f(C)^{\pi} \cap (D_1 \times \LL \times D_n)$ it
suffices to prove that $d \in D_1 \times \LL \times D_n$.  For each
variable $x_j$ lying inside of $X$ we have $d[x_j] = s_j \in D_j$.  In
turn, for each variable $x_j$ lying outside of $X$ its domain $D_j$
has two elements, so, by the assumption on ${\cal BASE}$, $D_j$ is the
same as the corresponding domain $D'_j$ of $f(C)^{\pi}$ and
consequently $d[x_j] \in D_j$, since $d \in D'_1 \times \LL \times
D'_n$.

So indeed $d \in C$ and hence $d[x_i] \neq a$ by the choice of $a$. This
proves validity of the equality rule $X = s \ra x_i \neq a$ for
$f(C)^{\pi}$. 

But $C$ is not closed under this rule since $a \in D_i$,
so $C$ is not rule consistent.
\end{proof}

\section{Membership Rule Consistency}
\label{sec:inclusion-rule-consistency}

In this section we consider computations that use exclusively
membership rules.  In the previous section we saw that the notion of
rule consistency is weaker than that of arc consistency for
constraints of arbitrary arity.  Here we show that by using the
membership rules we obtain a notion of local consistency that
coincides with arc consistency.


%
%
%

First, let us clarify the notion of a membership rule
by considering the following example.
\begin{example} \label{exa:waltz}
Consider a constraint on variables $x,y,z$, each with the
domain $\C{+,-, l, r}$, that is defined by the
following table:

\[
\begin{array}{|l|l|l|}
\hline
x & y & z \\ \hline \hline
+ & + & + \\
- & - & - \\
l & r & - \\
- & l & r \\
r & - & l \\ \hline
\end{array}
\]
This constraint is the so-called {\em fork\/} junction
in the language of \cite{waltz75}
for describing polyhedral scenes.
Note that the following three membership rules
\[
r_1 :=
x \in \C{+,-} \ra z \neq l,
\]
\[
r_2 := x \in \C{+} \ra z \neq l,
\]
and
\[
r_3 := x \in \C{-}, y \in \C{l} \ra z \neq l
\]
are all valid.  The membership rules $r_2$ and $r_3$
extend $r_1$ while the membership rule $r_1$ extends neither
$r_2$ nor $r_1$. Further,
the membership rules $r_2$ and $r_3$ are incomparable in the sense
that none extends the other.
\HB
\end{example}

Now, in analogy to Definition \ref{def:rule-cons}, we introduce the
following notion.

\begin{definition} \label{def:inclusion-rule-cons}
Consider a CSP ${\cal P}$ is based on a CSP ${\cal BASE}$.  Let
$C$ be a constraint of ${\cal P}$.
For some
constraint
$f(C)$ of  ${\cal BASE}$ and a permutation $\pi$, 
$C$ is based on $f(C)^{\pi}$.

\begin{itemize}

\item We call the constraint $C$ {\em membership rule consistent\/}
(w.r.t.  ${\cal BASE}$)
if it is closed under all membership rules that are valid for
$f(C)^{\pi}$.

\item We call a CSP {\em membership rule consistent\/} (w.r.t.  ${\cal
BASE}$)
if all its constraints are membership rule consistent.
\HB
\end{itemize}
\end{definition}

We now have the following result.
\begin{theorem} \label{thm:arccons}
Consider a CSP $\cal P$ based on a CSP ${\cal BASE}$.
Then $\cal P$ is arc consistent iff it is
membership rule consistent
w.r.t. ${\cal BASE}$.
\end{theorem}
\begin{proof}
($\Ra$) This part of the proof is a simple modification of the
proof of Theorem \ref{thm:rulecons}.

Assume that $\cal P$ is arc consistent. Choose a constraint $C$ of
$\cal P$ and consider a membership rule $X \in S \ra y \neq a$ that
is valid for 
$f(C)^{\pi}$, where $f$ and $\pi$ are as in Definition
\ref{def:based}.

Suppose by contradiction that $C$ is not closed under
this rule. So for $X := x_{1}, \LL, x_{k}$ and
$S := S_1, \LL, S_k$
the domain of each variable $x_{j}$ is included in $S_j$ and moreover
$a \in D$, where $D$ is the domain of the variable $y$.

By the arc consistency of $\cal P$ there exists $d \in C$ such that
$d[y] = a$.  Because of the form of the domains of the variables in
$X$, also $d[x_i] \in S_i$ for $i \in [1..k]$ holds.  Additionally,
because $\cal P$ is based on ${\cal BASE}$ we have $d \in
f(C)^{\pi}$.
But by assumption the rule $X \in S \ra y \neq a$ is valid for
$f(C)^{\pi}$, so
$d[y] \neq a$. A contradiction.
\vspace{1mm}

\NI
($\La$)
This part of the proof is a modification of the
proof of Theorem \ref{thm:rule-consistency}.

Suppose that some constraint $C$ of $\cal P$ is not arc consistent. We
prove that then $C$ is not membership rule consistent.  The constraint
$C$ is on some variables $x_1, \LL, x_n$ with respective domains $D_1,
\LL, D_n$.  For some $i \in [1..n]$ some $a \in D_i$ does not
participate in any solution to $C$.

Take $f(C)^{\pi}$, where $f$ and $\pi$ are as in Definition
\ref{def:inclusion-rule-cons}.  For appropriate domains 
$D'_1, \LL, D'_n$ of ${\cal
  BASE}$ we have $f(C)^{\pi} \sse D'_1 \times \LL \times D'_n$.

Let $D_{i_1}, \LL, D_{i_\ell}$ be the sequence of domains in $D_1,
\LL, D_{i-1}, D_{i+1}, \LL, D_n$ that are respectively different than
$D'_1, \LL, D'_{i-1}, D'_{i+1}, \LL, D'_n$.  Further, let $X :=
x_{i_1}, \LL, x_{i_\ell}$ and $S := D_{i_1}, \LL, D_{i_\ell}$.

Consider now the membership rule $X \in S \ra x_i \neq a$.  Take some $d \in
f(C)^{\pi}$ such that $d[x_{i_j}] \in D_{i_j}$ for $j \in [1.. \ell]$.
We show that $d \in C$.  Since $C = f(C)^{\pi} \cap (D_1 \times \LL
\times D_n)$ it suffices to prove that $d \in D_1 \times \LL \times
D_n$.  For each variable $x_j$ lying inside of $X$ we have $d[x_j] \in
D_j$.  In turn, for each variable $x_j$ lying outside of $X$ its
domain $D_j$ is the same as the corresponding domain $D'_j$ of
$f(C)^{\pi}$ in ${\cal BASE}$ and consequently $d[x_j] \in D_j$, since
$d \in D'_1 \times \LL \times D'_n$.

So indeed $d \in C$ and hence $d[x_i] \neq a$ by the choice of $a$.
This proves validity of the rule $X \in S \ra x_i \neq a$ for
$f(C)^{\pi}$.  But $C$ is not closed under this membership rule since
$a \in 
D_i$, so $C$ is not membership rule consistent.
\end{proof}

Example \ref{exa:not-arc-consistent} shows that the notions of rule
consistency and membership rule consistency do not coincide. To
see this difference better let us reconsider the CSP discussed
in this example.

We noted there that this CSP is not arc consistent and that it
is rule consistent. From the above theorem we know that this CSP
is not membership rule consistent. In fact, consider the following
membership rule:
\[
x \in \C{0,1} \ra y \neq 2.
\]

This membership rule is valid for the base constraint $C$ but the
restricted constraint $C \cap (D_1 \times D_2)$ is not closed under
this rule.  In conclusion, the membership rules are more powerful than
the equality rules.

%




As in Section \ref{sec:rule-consistency} we now provide an algorithm
that given a constraint generates the set of all minimal valid
membership rules.
We assume here that the considered constraint $C$ is defined on a
sequence of variables {\em VAR\/} of cardinality $n$.

Instead of assignments that are used in the {\sc Equality Rules
  Generation} algorithm we now need a slightly different notion. To
define it for each variable $x$ from {\em VAR\/} recall that we
denoted the set $\C{d[x] \mid d \in C}$ by $C[x]$.  By a {\em weak
  assignment\/} to a sequence of variables $X := x_1, \LL, x_k$ we
mean here a sequence $S_1, \LL, S_k$ of subsets of, respectively,
$C[x_1], \LL, C[x_k]$ such that some $d \in C$ exists such that
$d[x_i] \in S_i$ for each $i \in [1..k]$.

Intuitively, if we represent the constraint $C$ as a table with rows
corresponding to the elements of $C$ and the columns corresponding to
the variables of $C$ and we view each column as a set of elements,
then a weak assignment to $X$ is a tuple of subsets of the columns
that correspond to the variables of $X$ that ``shares'' an assignment.

In the algorithm below the weak assignments to a fixed sequence of
variables are considered in decreasing order in the sense that if the
weak assignments $S_1, \LL, S_k$ and $U_1, \LL, U_k$ are such that for
$i \in [1..k]$ we have $U_i \sse S_i$, then $S_1, \LL, S_k$ is
considered first.

\III

\NI
{\sc Membership Rules Generation} algorithm

\begin{alltt}
\emph{L := {\bf empty}};
FOR \emph{i:= 0} TO \emph{n-1} DO
   FOR each subset \emph{X} of \emph{VAR} of cardinality \emph{i} DO
      FOR each weak assignment \emph{S} to \emph{X} in decreasing order DO
         FOR each \emph{y} in \emph{VAR-X} DO
            FOR each element \emph{d} from the domain of \emph{y} DO
               \emph{r :=  X} \(\in\) \emph{S} \ra \emph{y} \(\neq\) \emph{d};
               IF \emph{r} is valid for \emph{C}
                  and it does not extend an element of \emph{L}
                  THEN \emph{{\bf insert}(r, L)}
               END
            END
         END
      END
   END
END
\end{alltt}
The test that one membership rule does not extend another can be 
implemented using Note \ref{not:rule-extends}.ii.

The following result establishes correctness of this algorithm.

\begin{theorem} \label{thm:irga}
Given a constraint $C$ the {\sc Membership Rules Generation} algorithm
produces
in {\it L} the set of all minimal valid membership rules for $C$.
\end{theorem}

\begin{proof}
  The proof is analogous to that of Theorem \ref{thm:rga}.  We only
  need to check that the membership rules are considered in such an
  order that if a rule $r_2$ extends a rule $r_1$, then $r_1$ is
  considered first.
This follows from directly from 
Note \ref{not:rule-extends}.ii.
\end{proof}

\section{Applications}
\label{sec:examples}
In this section we discuss the implementation of the
{\sc Equality Rules Generation} and {\sc Membership Rules Generation} algorithms
and discuss their use on selected domains.

\subsection{Constraint Handling Rules ({\tt CHR})}

In order to validate our approach we have realized in the Prolog
platform \eclipse{} a prototype implementation of both the
\textsc{Rules Generation} algorithm and the \textsc{Membership Rules
  Generation} algorithm.  We made a compromise between memory usage
and performance so that we could tackle some non-trivial problems (in
terms of size of the domains of variables, and in terms of arity of
constraints) in spite of the exponential complexity of the algorithms.
These implementations generate {\tt CHR} rules that deal with finite
domain variables using an \eclipse{} library.

Constraint Handling Rules ({\tt CHR}) of \cite{fruhwirth-constraint-95}
is a declarative language that
allows one to write guarded rules for rewriting constraints. These
rules are repeatedly applied until a fixpoint is reached. The rule
applications have a precedence over the usual resolution step of logic
programming.

A {\tt CHR} program is a finite set of {\tt CHR} rules. These rules
are basically of two types (there is a third type of rules which is a
combination of the first two types):  simplification rules and propagation rules.
When all guards are satisfied,
a simplification rule replaces constraints by simpler ones while
preserving logical equivalence, whereas a propagation rule adds
logically redundant constraints.
More precisely, these rules have the following form:

\[
\begin{array}{ll}
\textsf{simplification} &
H_1, \ldots, H_i <=> G_1, \ldots, G_j ~|~ B_1, \ldots, B_k\\
\textsf{propagation} &
H_1, \ldots, H_i ==> G_1, \ldots, G_j ~|~ B_1, \ldots, B_k
\end{array}
\]
where 
\begin{itemize}
\item $i > 0$, $j \geq 0$, $k\geq 0$,

\item the multi-head $H_1, \ldots, H_i$ is a non-empty sequence of CHR constraints, 

\item the guard $G_1, \ldots,
G_j$ is a sequence of built-in constraints,

\item the body 
$B_1, \ldots, B_k$ is a sequence of built-in and CHR constraints.
\end{itemize}

Our equality rules and membership rules can be modelled by means of
propagation 
rules. To illustrate this point consider some constraint $cons$
on three variables, $A,B,C$, each with the domain $\C{0,1,2}$.

The \textsc{Rules Generation} algorithm generates rules such as
$(A,C)=(0,1) \rightarrow B \neq 2$. This rule is translated into a
{\tt CHR}
rule of the form: \verb+cons(0,B,1)+ \verb+==>+ \verb+B##2+.  Now, when
a constraint
in the program query unifies with \verb+cons(0,B,1)+, this rule is
fired and the value 2 is removed from the domain of the variable
\verb+B+.

In turn, the \textsc{Membership Rules Generation} algorithm generates
rules such as $(A,C) \in (\{0\},\{1,2\}) \rightarrow B \neq 2$.
This rule is translated into the {\tt CHR} rule
\begin{verbatim}
cons(0,B,C) ==>in(C,[1,2]) | B##2
\end{verbatim}
where the {\tt in} predicate is defined by
\begin{verbatim}
in(X,L):- dom(X,D), subset(D,L).
\end{verbatim}
So {\tt in(X,L)} holds if the current domain of the variable {\tt
X} (yielded by the built-in {\tt dom} of \eclipse{}) is included in
the list {\tt L}.

Now, when a constraint unifies with \verb+cons(0,B,C)+ {\em and\/} the
current domain of the variable \verb+C+ is included in \verb+[1,2]+,
the value 2 is removed from the domain of \verb+B+.
So for both types of rules we achieve the desired effect.

In the examples below we combine the rules with the same premise into
one rule in an obvious way and present these rules in the {\tt CHR}
syntax.

\subsection{Generating the rules}
\label{subsec:generating}

We begin by discussing the generation of equality rules and membership rules
for some selected domains.
The times given refer to an implementation ran on a Silicon Graphics O2
with 64 Mbytes of memory and a 180 MHZ processor.

\paragraph{Boolean constraints}
As the first example consider the Boolean constraints, for
example the conjunction constraint {\tt and(X,Y,Z)}
of Example \ref{exa:conjunction}.
The {\sc Equality Rules Generation} algorithm
generated in 0.02 seconds the following six equality rules:
\begin{verbatim}
and(1,1,X) ==> X##0.
and(X,0,Y) ==> Y##1.
and(0,X,Y) ==> Y##1.
and(X,Y,1) ==> X##0,Y##0.
and(1,X,0) ==> X##1.
and(X,1,0) ==> X##1.
\end{verbatim}

Because the domains are here binary we can
replace the conclusions of the form {\tt U \#\# 0} by {\tt U = 1}
and {\tt U \#\# 1} by {\tt U = 0}.
These rules are somewhat different than the well-known rules that can be found e.g. in
\cite{FruehwirthJLP98}[page 113], where instead of the rules

\begin{verbatim}
and(1,1,X) ==> X##0.
and(1,X,0) ==> X##1.
and(X,1,0) ==> X##1.
\end{verbatim}
the rules
\begin{verbatim}
and(1,X,Y) ==> X = Y.
and(X,1,Y) ==> X = Y.
and(X,Y,Z), X = Y ==> Y = Z.
\end{verbatim}
appear.
We shall discuss this matter in Section \ref{sec:related}.

In this case, by virtue of Theorem \ref{thm:rule-consistency}, the
notions of rule and arc consistency coincide, so the above six
equality rules
characterize the arc consistency of the {\tt and} constraint. Our
implementations of the {\sc Equality Rules Generation} and the {\sc Membership
Rules Generation} algorithms yield here the same rules.

\paragraph{Three valued logic}

Next,  consider the three valued logic of
\cite{Kle52}[page 334] that consists of three values,
t (true), f (false) and u (unknown).
We only consider here the crucial equivalence relation $\equiv$ defined
by
the truth table

\[
\begin{array}{|c|ccc|} \hline
\equiv & $t$ & $f$ & $u$ \\ \hline
$t$ & $t$ & $f$ & $u$ \\
$f$ & $f$ & $t$ & $u$ \\
$u$ & $u$ & $u$ & $u$ \\
\hline
\end{array}
\]
that determines a ternary constraint with nine triples.
We obtain for it 20 equality rules and 26 membership rules.
Typical examples are
\begin{verbatim}
equiv(X,Y,f) ==> X##u,Y##u.
\end{verbatim}
and
\begin{verbatim}
equiv(t,X,Y) ==> in(Y,[f, u]) | X##t.
\end{verbatim}

\paragraph{Six valued logic}

In \cite{vanhentenryck-constraint-using} the constraint logic
programming language CHIP is used for the automatic test-pattern
generation (ATPG) for the digital circuits.  To this end the authors
define a specific six valued logic and provide some rules (expressed
in the form of so-called demons) to carry out the constraint
propagation.  The {\tt and6} constraint in question is defined by means
of the following table:
\[
\begin{array}{|c|cccccc|} \hline
{\tt and6} & 0 & 1 & $d$ & $dnot$ & $e$ & $enot$ \\ \hline
0 & 0 & 0 & -- & -- & 0 & 0 \\
1 & 0 & 1 & $d$ & $dnot$ & $e$ & $enot$ \\
$d$ & -- & $d$ & -- & -- & $d$ & -- \\
$dnot$ & -- & $dnot$ & -- & -- & --  & $dnot$ \\
$e$ & 0 & $e$ & $d$ & -- & $e$ & 0 \\
$enot$ & 0 & $enot$ & -- & $dnot$ & 0 & $enot$ \\
\hline
\end{array}
\]

The {\sc Equality Rules Generation} algorithm generated 41 equality rules in
0.15 seconds, while the {\sc Membership Rules Generation} algorithm
generated 155 membership rules in 14.35 seconds.  
The generated rules enforce, respectively,
rule consistency and arc consistency, while it is not clear
what notion of local consistency is enforced by the (valid) rules of
\cite{vanhentenryck-constraint-using}[page 133] because some of the
latter ones allow equalities between the variables in the premise.
This makes the comparison in terms of strength of the entailed
notion of local consistency difficult.
It is clear that our approach is more systematic and fully automatic.
(In fact, we found 
two typo's in the rules of \cite{vanhentenryck-constraint-using}[page 133].)

\paragraph{Propagating signs}

As a next example consider the rules for propagating signs
in arithmetic expressions, see, e.g.,
\cite{davis87}[page 303]. We limit ourselves to the
case of multiplication.
Consider the following table:

\[
\begin{array}{|r|llll|} \hline
\times & $neg$  & $zero$  & $pos$   & $unk$ \\
\hline
$neg$     & $pos$   & $zero$  & $neg$   & $unk$ \\
$zero$    & $zero$  & $zero$  & $zero$  & $zero$ \\
$pos$     & $neg$   & $zero$  & $pos$   & $unk$ \\
$unk$     & $unk$   & $zero$  & $unk$   & $unk$ \\
\hline
\end{array}
\]
This table determines a ternary constraint {\tt msign}
that consists of 16 triples, for
instance (neg, neg, pos) that denotes the fact that the
multiplication of two negative numbers yields a positive number.
The value ``unk'' stands for ``unknown''.
The {\sc Equality Rules Generation} algorithm
generated in 0.08 seconds 34 equality rules.
A typical example is \verb+msign(X,zero,Y) ==> Y##pos,Y##neg,Y##unk+.

In turn, the {\sc Membership Rules Generation} algorithm
generated in 0.6 seconds 54 membership rules. A typical example is
\begin{verbatim}
msign(X,unk,Y) ==> in(Y,[neg, pos, zero]) | X##pos,X##neg
\end{verbatim}
that corresponds to the following two membership rules for the
constraint {\tt 
msign(X,Z,Y)}:
\[
(Z,Y) \in (\C{{\tt unk}}, \C{{\tt neg, pos, zero}}) \ra X \neq {\tt pos}
\]
and
\[
(Z,Y) \in (\C{{\tt unk}}, \C{{\tt neg, pos, zero}}) \ra X \neq {\tt neg}.
\]

\paragraph{Waltz' language for describing polyhedral scenes}

Waltz' language consists of four constraints. One of them, the fork
junction was already mentioned in Example \ref{exa:waltz}.  The {\sc
  Equality Rules Generation} algorithm generated for it 12 equality
rules and the {\sc Membership Rules Generation} algorithm 24 membership
rules.

Another constraint, the so-called T junction, is defined by
the following table:

\[
\begin{array}{|c|c|c|}
\hline
x & y & z \\ \hline \hline
{\tt r} & {\tt l} & {\tt +} \\
{\tt r} & {\tt l} & {\tt -} \\
{\tt r} & {\tt l} & {\tt r} \\
{\tt r} & {\tt l} & {\tt l} \\ \hline
\end{array}
\]

In this case the {\sc Equality Rules Generation} algorithm and the {\sc
  Membership Rules Generation} algorithm both generate the same output
that consists of just one rule:
\begin{verbatim}
t(X,Y,Z) ==> X##'l',X##'-',X##'+',Y##'r',Y##'-',Y##'+'.
\end{verbatim}
So this rule characterizes both rule consistency
and arc consistency for the CSP's based on the T junction.

For the other two constraints, the L junction and the arrow junction,
the generation of the equality rules and membership rules is equally
straightforward.

\subsection{Using the rules}

Next, we show by means of some examples how the generated rules can be
used to reduce or to solve specific queries. Also, we show how using
compound constraints we can achieve local consistency notions
that are stronger than arc consistency for constraints of arbitrary
arity.

\paragraph{Waltz' language for describing polyhedral scenes}

The following predicate describes the impossible scene given in
Figure \ref{fig:impossible} and taken from 
\cite{Win92}[page 262]:

\begin{figure}[htbp]
  \begin{center}
\makebox[\textwidth]{\psfig{figure=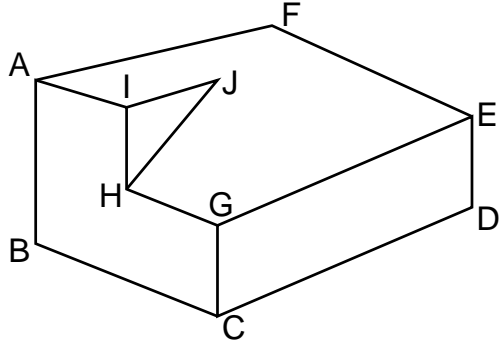,height=4.5cm}}
\caption{\label{fig:impossible} Impossible scene}
  \end{center}
\end{figure}

\begin{verbatim}
imp(AF,AI,AB,IJ,IH,JH,GH,GC,GE,EF,ED,CD,CB):-
       S1=[AF,AI,AB,IJ,IH,JH,GH,GC,GE,EF,ED,CD,CB],
       S2=[FA,IA,BA,JI,HI,HJ,HG,CG,EG,FE,DE,DC,BC],
       append(S1,S2,S), S :: [+,-,l,r],

       arrow(AF,AB,AI), l(BC,BA), arrow(CB,CD,CG), l(DE,DC),
       arrow(ED,EG,EF), l(FA,FE), fork(GH,GC,GE), arrow(HG,HI,HJ),
       fork(IA,IJ,IH), l(JH,JI),

       line(AF,FA), line(AB,BA), line(AI,IA), line(IJ,JI),
       line(IH,HI), line(JH,HJ), line(GH,HG), line(FE,EF),
       line(GE,EG), line(GC,CG), line(DC,CD), line(ED,DE),
       line(BC,CB).
\end{verbatim}
where the supplementary constraint {\tt line} is defined by the
following table:

\[
\begin{array}{|c|c|}
\hline
$x$ & $y$  \\ \hline \hline
{\tt +} & {\tt +} \\
{\tt -} & {\tt -} \\
{\tt l} & {\tt r} \\
{\tt r} & {\tt l} \\ \hline
\end{array}
\]

Here and elsewhere we use the \eclipse{} built-in {\tt ::} to declare
variable domains.

When using the equality rules obtained by the {\sc Equality Rules Generation}
algorithm and associated with the \verb+fork+, \verb+arrow+, \verb+t+,
\verb+l+, and \verb+line+ constraints, the query
\begin{verbatim}
imp(AF,AI,AB,IJ,IH,JH,GH,GC,GE,EF,ED,CD,CB)
\end{verbatim}
reduces in 0.009 seconds the variable domains to
\texttt{AF $\in$ [+,-, l]},
\texttt{AI $\in$ [+,-]},
\texttt{AB $\in$ [+,-,r]},
\texttt{IJ $\in$ [+,-,l,r]},
\texttt{IH $\in$ [+,-,l,r]},
\texttt{JH $\in$ [+,-,l,r]}, \\
\texttt{GH $\in$ [+,-,l,r]},
\texttt{GC $\in$ [+,-,l,r]},
\texttt{GE $\in$ [+,-,l,r]},
\texttt{EF $\in$ [+,-]},\\
\texttt{ED $\in$ [+,-,l]},
\texttt{CD $\in$ [+,-,r],} and
\texttt{CB $\in$ [+,-,l]}.

But
some constraints remain unsolved, so we need to add a labeling
mechanism to prove the inconsistency of the problem.  On the other
hand, when using the membership rules, the inconsistency is detected
without any labeling in 0.06 seconds.

In the well-known example of the cube given in Figure~12.15 of
\cite{Win92}[page 260] the membership rules are also more powerful
than the equality rules and both sets of rules reduce the problem but in both
cases labeling is needed to produce all four solutions.

Comparing the constraint solver based on the membership rules to the
constraint solver based on the equality rules is not easy: although
propagation is
more efficient with the membership rules, the solver based on the equality 
rules can
sometimes be faster depending on the structure of the problem and
on whether the labeling is needed.

We also compared the solvers generated by the implementations of
the \textsc{Equality Rules Generation} and \textsc{Membership Rules Generation}
algorithms to
the approach described in~\cite{By97} and based on
meta-programming in Prolog. We ran the same examples and drew the
following conclusions.  For small examples our
solvers were less efficient than the ones of By (with factors varying
from 2 to 10).
However, for more complex examples, our solvers became
significantly more efficient, with factors varying from 10 to 500.
This can be attributed to the increased role of the constraint
propagation
that reduces backtracking and that is absent in By's approach.

\paragraph{Temporal reasoning}

In \cite{All83}'s approach to temporal reasoning the entities are
intervals and the relations are temporal binary relations between
them.  \cite{All83} found that there are 13 possible temporal
relations between a pair of events, namely {\tt before, during,
  overlaps, meets, starts, finishes}, the symmetric relations of these
six relations and {\tt equal}. We denote these 13 relations respectively
by {\tt  b,d,o,m,s,f,}{\tt b-,d-,o-,m-,s-,f-,e} and their set by {\tt TEMP}.

Consider now three events, {\tt A, B} and {\tt C} and suppose that we
know the temporal relations between the pairs {\tt A} and {\tt B}, and
{\tt B} and {\tt C}. The question is what is the temporal relation
between {\tt A} and {\tt C}.  To answer it \cite{All83} provided
a 13 $\times$ 13 table.  This table determines a ternary constraint
between a triple of events, {\tt A, B} and {\tt C} that we denote by
{\tt allen}.  For example,

\[
{\tt (overlaps, before, before)} \in {\tt allen}
\]
since {\tt A} {\tt overlaps\/}
{\tt B} and {\tt B} is {\tt before\/} {\tt C}
implies that  {\tt A} is {\tt before\/} {\tt C}.

Using this table, the {\sc Equality Rules Generation}
algorithm produced for the constraint {\tt allen} 498 equality rules in 31.16
seconds. 
In contrast, we were unable to generate all membership rules in less than
24 hours. This shows the limitations of our implementation.
We tried the generated set of equality rules to solve the following problem
from~\cite{All83}: ``John was not in the room when I touched the
switch to turn on the light.''. We have here three events: {\tt S}, the
time of touching the switch; {\tt L}, the time the light was on; and
{\tt J},
the time that John was in the room.  Further, we have two relations:
{\tt R1} between {\tt L} and {\tt S}, and {\tt R2} between {\tt S} and
{\tt J}. This problem
is translated into the CSP  $\p{{\tt allen}}{{\tt R1} \in {\tt [o-,m-]},
{\tt  R2} \in {\tt [b,m,b-,m-]}, {\tt R3} \in {\tt TEMP}}$,
where {\tt allen} is the above constraint
on the variables {\tt R1}, {\tt R2}, {\tt R3}.

To infer the relation {\tt R3} between {\tt L} and {\tt J} we can use
the
following query\,\footnote{Since no variable is instantiated, we need
  to perform labeling to effectively apply the rules.}:
\begin{verbatim}
  R1::[o-,m-],
  R2::[b,m,b-,m-],
  R3::[b,d,o,m,s,f,b-,d-,o-,m-,s-,f-,e],
  allen(R1,R2,R3),
  labeling([R1,R2,R3]).
\end{verbatim}
We then obtain the following solutions in 0.06 seconds:\\
\texttt{(R1,R2,R3)} $\in$
\texttt{\{(m-,b,b),}
\texttt{(m-,b,d-),}
\texttt{(m-,b,f-),}
\texttt{(m-,b,m),}
\texttt{(m-,b,o),}\\
\texttt{(m-,b-,b-),}
\texttt{(m-,m,e),}
\texttt{(m-,m,s),}
\texttt{(m-,m,s-),}
\texttt{(m-,m-,b-),}
\texttt{(o-,b,b),}\\
\texttt{(o-,b,d-),}
\texttt{(o-,b,f-),}
\texttt{(o-,b,m),}
\texttt{(o-,b,o),}
\texttt{(o-,b-,b-),}
\texttt{(o-,m,d-),}\\
\texttt{(o-,m,f-),}
\texttt{(o-,m,o),}
\texttt{(o-,m-,b-)\}}.

To carry on (as in~\cite{All83}), we now complete the problem
with: ``But John was in the room later while the light went out.''.
This is translated into: ``{\tt L} {\tt overlaps}, {\tt starts}, or
{\tt is during} {\tt J}'',
i.e., {\tt R3} $\in$ {\tt [o,s,d]}.

We now run the following query:
\begin{verbatim}
  R1::[o-,m-],
  R2::[b,m,b-,m-],
  R3::[o,s,d],
  allen(R1,R2,R3),
  labeling([R1,R2,R3]).
\end{verbatim}
and obtain four solutions in 0.04 seconds:
\texttt{(R1,R2,R3)} $\in$
\texttt{\{(m-,b,o),}
\texttt{(m-,m,s),}
\texttt{(o-,b,o),}
\texttt{(o-,m,o)\}}.

\paragraph{Three valued logic}
Next, consider the \texttt{and3} constraint in the three valued logic of
\cite{Kle52}[page 334] represented by the truth table

\[
\begin{array}{|c|ccc|} \hline
\texttt{and3}   & $t$ & $f$ & $u$ \\ \hline
$t$            & $t$ & $f$ & $u$ \\
$f$            & $f$ & $f$ & $f$ \\
$u$            &  $u$ & $f$ & $u$ \\
\hline
\end{array}
\]  

Typical examples of the 16 generated equality rules and of the 18 generated membership rules are:
\begin{verbatim}
and3(u,u,X) ==> X##0.
\end{verbatim}
and 
\begin{verbatim}
and3(X,Y,Z) ==> in(X,[0, u]) | Z##1.
\end{verbatim}
Consider now the query:
\begin{verbatim}
[X,Y,Z,T,U]::[0,1,u], and3(X,Y,Z), and3(T,U,Z), Z##0,Y##u, X##u.
\end{verbatim}
Using the membership rules we reach the answer in the form of 
a complete assignment to all
the variables, namely \verb+X = 1,Y = 1,Z = 1,T = 1,U = 1+, whereas
using only the equality rules we do not obtain any reduction and
labeling is needed to produce the same answer.

\paragraph{Full adder}
It is often the case that dealing with a compound constraint directly
yields a stronger notion of local consistency than when dealing with
each of its constituents separately. A prototypical example is the
well-known \texttt{alldifferent} constraint on $n$ variables that can
be decomposed into $\frac{n \cdot (n-1)}{2}$ binary disequality
constraints. Then arc consistency enforced on \texttt{alldifferent} is
stronger than arc consistency enforced on each of the disequality
constraints separately.

The same phenomenon arises for rule consistency.
We illustrate it by means of 
the already discussed in Example \ref{exa:full-adder}
full adder circuit. It can be defined by the following
constraint logic program (see, e.g., \cite{FruehwirthJLP98})
that uses the Boolean constraints
{\tt and}, {\tt xor} and {\tt or}:
\begin{verbatim}
add(I1,I2,I3,O1,O2):-
        [I1,I2,I3,O1,O2,A1,A2,X1]:: 0..1,
        xor(I1,I2,X1),
        and(I1,I2,A1),
        xor(X1,I3,O2),
        and(I3,X1,A2),
        or(A1,A2,O1).
\end{verbatim}
The query \verb+add(I1,I2,I3,O1,O2)+ followed by a labeling mechanism
generates the explicit definition (truth table) of the {\tt
  full\_adder} constraint with eight entries such as
\begin{verbatim}
full_adder(1,0,1,1,0).
\end{verbatim}


We can now generate the equality rules for the compound 
constraint (here the {\tt full\_adder} constraint) that is defined by
means of some basic constraints (here the {\tt and, or} and {\tt xor}
constraints).  These rules refer to the compound constraint
and allow us to reason about it directly instead of by using the rules
that deal with the basic constraints.

In the case of the {\tt full\_adder} constraint the {\sc Equality Rules
  Generation} algorithm generated 52 equality rules in 0.27 seconds.  The
constraint propagation carried out by means of these equality rules is more
powerful than the one carried out by means of the equality rules generated for
the {\tt and, or} and {\tt xor} constraints.
For example,
the query \verb+[X,Y,Z]::[0,1], full_adder(1,X,Y,Z,0)+ reduces {\tt Z}
to
1 whereas the query
\verb+[X,Y,Z]::[0,1], add(1,X,Y,Z,0)+  does not reduce {\tt Z} at all.

So rule consistency for a compound constraint defined
by means of the basic constraints is indeed in general stronger than the rule
consistency for the basic constraints treated separately.  In fact, in
the above case the equality 
rules for the {\tt full\_adder} constraint yield the relational
(1,5)-consistency notion of \cite{DvB97}, whereas by virtue of
Theorem \ref{thm:rule-consistency}, the equality rules for the {\tt and, or}
and {\tt xor} constraints yield a
weaker notion of arc consistency.

\section{Related Work}
\label{sec:related}

\subsection{Relation Between Constraint Programming and Rule-Based Programming}

In a number of papers a link was made between constraint programming
and rule-based programming.  To start with, in \cite{MR91} a general
study of constraint propagation was undertaken by defining the notion
of a relaxation rule and by proposing a general relaxation algorithm
that implements constraint propagation by means of a repeated
application of the relaxation rules.  However, this abstract view of
constraint programming cannot be realized in a simple way since the
application of a relaxation rule is a complex process.

In \cite{Apt98a} we showed how constraint programming can be couched
in proof theoretic terms by viewing the programming process as the
task of proving the original CSP. In the proposed framework two types
of rules were proposed: deterministic ones and the splitting ones.
Further, the deterministic rules were either concerned with domain
reduction or constraint reduction. In the former case the rules were
called domain reduction rules and in the latter case constraint
reduction rules.  Such rules are high-level abstractions and on the
implementation level they can involve complex computations.

It is useful to see that the rule-based approach to constraint
programming proposed in this paper is an instance of this proof
theoretic view of constraint programming. Namely, the equality rules
and the membership rules are examples of the domain reduction rules
while labeling, the formal treatment of which is omitted here, is an
example of a splitting rule.

The important gain is that the implementation of the considered here
equality rules and membership rules boils down to a straightforward
translation of them into the {\tt CHR} syntax. This leads to an
implementation of this approach to constraint programming by means of
constraint logic programming.  The important limitation is that this
approach applies only to the CSP's built out of predefined, explicitly
given finite constraints.

A similar approach to constraint programming to that of \cite{Apt98a}
was proposed in \cite{Cas98}.  In his approach the proof rules are
represented as rewrite rules in the already mentioned in the
introduction programming language {\sf ELAN}.  
The rules use a richer syntax than here considered by referring to
arbitrary constraints and to expressions of the form $x \in D$, where
$D$ is the current domain of the variable $x$. In particular no
constraint specific rules were considered.  Instead, the emphasis
was on showing how the general techniques of constraint programming,
in particular various search strategies, can be expressed in the form
of rules.

\subsection{Generation of Rules}

Let us turn now to an overview of the recent work on rules generation.
Building upon the work presented in~\cite{AM99} two articles appeared
in which algorithms were presented that aim at improving the
expressivity of the rules and at a more economic representation.

In~\cite{RingeissenMonfroy:LNAI:2000} rules similar to equality rules
were considered. The most significant improvement is the use of
parameters (i.e., unspecified constants) that leads to a decrease in
the number of generated rules.  Parameters are also a means for
deducing equalities of variables in the right-hand side of rules.  For
instance, consider the following two rules with parameters $a_1$ and
$a_2$ for a constraint $C$ taken from~\cite{RingeissenMonfroy:LNAI:2000}:

  \begin{equation}
x_1=a_1 \ra x_2=a_1 \wedge x_3=1 \label{d1}
  \end{equation}

  \begin{equation}
x_2=a_2 \ra  x_1=a_2 \wedge x_3=1 \label{d2}
  \end{equation}
Rule~(\ref{d1}) means that whatever the value of $x_1$ is, $x_2$ is
equal to $x_1$ and $x_3$ is equal to 1. From rules~(\ref{d1}) and
(\ref{d2}) an equality between the variables on the right-hand side of
the rule can be deduced (note that the resulting rule can always be applied):
\[
\ra x_2=x_1 \wedge x_3=1
\]

To generate such rules with parameters
\cite{RingeissenMonfroy:LNAI:2000} combine unification in finite
algebra with a rule generation algorithm. The size of the generated
set of rules significantly depends on an ordering on variables. 
It still needs to be clarified what is the counterpart of the
notion of a minimal rule in this framework and whether the generated
rules with parameters enforce rule-consistency.
\medskip

Sets of rules generated in~\cite{Abdennadher:Rigotti:CP00} are even
more compact and more expressive: multiple occurrences of variables and
conjunction of constraints with shared variables are allowed in the
left-hand side of rules.  Moreover, the user has the
possibility to specify the admissible syntactic form of the rules:
more specifically, right-hand sides of rules can
consists of more complex constraints than (dis)-equality constraints.
Here are two examples of rules (taken from two different sets of
rules of~\cite{Abdennadher:Rigotti:CP00}):
\begin{eqnarray}
and(x,x,z) \ra x=z. \label{dd1}\\
and(x,y,z), neg(x,y) \ra z=0. \label{dd2}
\end{eqnarray}
In rule~(\ref{dd1}) equality between variables is deduced using a double occurrence
of the variable $x$ in the head, and rule~(\ref{dd2}) defines interaction
between two constraints, $and$ and $neg$.

\cite{Abdennadher:Rigotti:CP00} also investigated what form of local
consistency is enforced by the rules generated by their algorithms.
When using the given constraint together with equality (between a
variable and a value) on the left-hand side, and only disequality
(between a variable and a value) on the right-hand side of the rules,
the generated rules enforce rule consistency.  However, in general,
the enforced local consistency is stronger than rule consistency.
In particular, it is plausible that membership rule consistency (i.e., arc
consistency) is enforced when disequality constraints are allowed both
on the right- and left-hand sides of rules.  
\bigskip

In \cite{Apt00b} two sets of rules for Boolean constraints were
compared.  One of them is the one presented in Subsection
\ref{subsec:generating} and the other the already mentioned set of
rules from \cite{FruehwirthJLP98}[page 113] (with one difference
irrelevant for the subsequent discussion).  While both sets of rules
enforce the arc consistency, it turns out that they are not
equivalent.  In fact, if a Boolean CSP with non-empty domains is closed under the
rules from the second set, then it is closed under the rules from the first set.
The converse does not hold since the CSP $\p{x \A y = z}{x \in \C{1}, y \in \C{0,1}, z
  \in \C{0,1}}$ is closed under the first set of rules but not under
the second one, since it is not closed under the
rule {\tt and(1,X,Y) ==> X = Y}.

\subsection{Local Consistency Notions}


In this paper we considered local consistency by focusing on
individual, arbitrary, non-binary, constraints. In the literature
algorithms for achieving such notions of local consistency usually
concentrated on achieving arc consistency of non-binary CSP's. Among
them the algorithm \texttt{CN} of~\cite{Mackworth:IJCAI77}, and
\texttt{GAC4} of~\cite{MM88} are respectively based on ideas similar
to the AC-3 and AC-4 algorithms for binary constraints. However, in practice
\texttt{CN} is usable only for ternary constraints and small domains,
and has a large worst-case time complexity. On the other hand, the
large space complexity of \texttt{GAC4} makes it usable only for
constraints of small size.

The \texttt{GAC-schema} of~\cite{Bessiere:Regin:IJCAI97} was designed
to enforce arc consistency on non-binary constraints while keeping a
reasonable time and space complexity. It is based on an AC-7 like
schema and allows constraints to be given explicitly, either in a
positive way as in our case, or in a negative way, i.e., in the form of ``forbidden''
tuples, or implicitly in the form of predicates. 
In order to make our framework as general as the \texttt{GAC-schema},
we could think of generating the allowed tuples by testing all possible
tuples. However, this would almost always be impractical because of space
considerations. 

In ~\cite{Bessiere:CP99} the following opinion was voiced on
non-binary constraints: ``Perhaps we should accept the idea that the
constraint solving tool of the next years will apply different levels
of local consistency on different constraints at each node of the
search tree, ...''. Our framework is amenable to such a view since we
can generate rules for enforcing various types of local consistencies.
For example, we can generate some equality rules for some constraints
and some membership rules for other constraints, and then apply only
the resulting set of rules.

Another consideration is that our constraint solving process consists
of two separate phases: first a generation of the rules (a sort of
compilation of the truth table of a constraint) which is done once and
for all, followed by the application of the rules. Thus, the set of
generated rules can be modified during the application phase, for
example by combining some rules (for a more efficient application of the
rules), by removing some rules or by strengthening some conditions to
weaken the domain reduction process.  
\medskip

While preparing this revised version of the paper we noted that a
similar notion to our rule consistency notion was introduced in the
context of the theory of fuzzy sets, see \cite{PG98}[pages 252-261].
The notion there considered deals with rules of the form ``if $x$ is
$A$, then $y$ is $B$'', where $A$ and $B$ are fuzzy sets. In spite of
the same name used (namely, rule consistency), the uses of both
notions are different. In our case, we employ it to reduce a specific
CSP to a smaller one that is rule consistent but can be
inconsistent, while in the case of the fuzzy set theory the corresponding
notion is used to detect conditions for ``potential inconsistency''
that arises when the rules express contradictory knowledge.
\medskip

Finally, let us mention that rule generation appears in other areas of
computer science. Typical examples are: programs for machine learning
that construct a model of the knowledge using decision
trees and production rules (see, e.g., \cite{Qui93})
inductive logic programming, a logic-based approach to machine
learning where logic programming rules are inferred from positive and
negative examples and a background knowledge (see, e.g.,
\cite{MR94}), and data mining that aims at extracting high-level
representations in the form of patterns and models from data
(see, e.g., \cite{AMSTV96} where so-called association rules
are generated.).

\section{Conclusions}
\label{sec:conclusions}

The aim of this paper was to provide a framework in which constraint
programming can be entirely reduced to rule-based programming.  It
involved constraint satisfaction problems built out of explicitly
given constraints. In the case the latter constraints are 
 defined over small finite domains these CSP's can be often
solved by means of automatically generated constraint propagation
algorithms.

We argued that such CSP's often arise in practice and consequently the
methods here developed can be of practical use.  We believe that the
approach of this paper could be applied to a study of various decision
problems concerning specific multi-valued logics and this in turn
could be used for an analysis of digital circuits (see, e.g.,
\cite{Mut76} where a nine valued logic is used).  Other applications
could involve non-linear constraints over small finite
domains and the analysis of polyhedral scenes in presence of shadows
(see \cite{waltz75}).

The introduced notion of rule consistency is weaker than arc
consistency and can be in some circumstances the more appropriate one
to use.  For example, for the case of temporal reasoning considered in
the last section we easily generated all 498 equality rules that
enforce rule consistency whereas 24 hours turned out not be enough to
generate the membership rules that enforce arc consistency.  (For a
more precise summary of the tests carried out see the appendix.)

In this paper, we focused on systematic and automated aspects of
rule-based constraint solvers. At present stage it is difficult to
compare the performance of our
method (based on rule generation and subsequent rule application) with
other methods based on classical constraint propagation algorithms.
The reason is that our approach is currently implemented by means of
{\tt CHR} rules that are applied on top of Prolog while the built-in
constraint propagation algorithms are usually implemented at a lower
level.

Finally, the notions of rule consistency and membership rule
consistency could be parametrized by the desired maximal number of
variables used in the rule premises. Such parametrized versions of
these notions could be useful when dealing with constraints involving
a large number of variables.  Both the {\sc Equality Rules Generation}
algorithm and the {\sc Membership Rules Generation} algorithm and their
implementations can be trivially adapted to such parametrized notions.

The approach proposed in this paper could be easily integrated into
constraint logic programming systems such as \eclipse{}. This could be
done by providing an automatic constraint propagation by means of the
equality rules or the membership rules for flagged predicates that are
defined 
by a list of ground facts, much in the same way as now constraint
propagation for linear constraints over finite systems is
automatically provided.

\section*{Acknowledgements}
We would like to thank Thom Fr\"{u}hwirth, Andrea Schaerf and all
three referees for several useful suggestions.
Victor Marek provided material for Section \ref{sec:example}
and suggested Section \ref{sec:outcomes}.




\section*{Appendix}

\begin{table}[htbp]
\[
\begin{array}{||l||c|c|c||c|c||c|c||}\hline
{\rm Constraint} & {\rm Arity}    & {\rm Domain}  & {\rm Cardinality} & 
     {\rm Equality} & {\rm Gen.}     & {\rm Member.} & {\rm Gen.} \\
                 &                & {\rm Size}       &                &
     {\rm Rules} & {\rm (in~s.)}  & {\rm Rules} & {\rm (in~s.)}\\ \hline\hline
fork      & 3 & 4  & 5   & 12  & 0.05  & 24   & 0.65 \\
t         & 3 & 4  & 4   & 1   & 0.02  &  1   & 0.07 \\ \hline

not        & 2 & 2  & 2   & 4   & 0.01  &  4   & 0.19 \\
not_3      & 2 & 3  & 3   & 6   & 0.02  &  6   & 0.29 \\
not_4      & 2 & 4  & 4   & 8   & 0.02  &  8   & 0.5  \\
not_6      & 2 & 6  & 6   & 12  & 0.03  &  12  & 0.14 \\
not_8      & 2 & 8  & 8   & 16  & 0.05  &  16  & 0.67 \\
not_9      & 2 & 9  & 9   & 18  & 0.07  &  18  & 1.57 \\ \hline

and        & 3 & 2  & 4   & 6   & 0.02  &  6   & 0.08 \\
and_3      & 3 & 3  & 9   & 16  & 0.04  & 18   & 0.13 \\
and_4      & 3 & 4  & 16  & 26  & 0.08  & 43   & 0.6 \\
and_6      & 3 & 6  & 24  & 41  & 0.15  & 155  & 14.35 \\
and_8      & 3 & 8  & 64  & 96  & 0.57  & 622  & 351.16 \\
and_9      & 3 & 9  & 81  & 134 & 1.07  & 1294 & 1777 \\ \hline

msign     & 3 & 4  & 16  & 34  & 0.08  & 54   & 0.6 \\
fulladder & 5 & 2  & 8  & 52  & 0.29  & 52   & 0.38 \\
b10m      & 4 & 10 & 100 & 362 & 14.83 & -   & - \\
allen     & 3 & 13 & 409 & 498 & 31.16 & -   & - \\ \hline
\end{array}
\]
\caption{\label{tests}  Generation of equality and membership rules}
\end{table}

Table~\ref{tests} illustrates the generation of the
equality rules and the membership rules
for various natural constraints. 
The first column gives the name of the constraint,
the second its arity, the third the cardinality of the
domains of variables, and the fourth the cardinality of the constraint.
The subsequent two columns show the outcome of the {\sc
Equality Rule Generation} algorithm: first the number of equality
rules generated, and then the computation time in seconds; the last
two columns provide this information for the membership rules.

The {\em fork, t, msign, allen}, and {\em fulladder\/} constraints represent the
previously described constraints. The $not_i$ and $and_i$ constraints,
where $i \in \{3,4,6,8,9\}$, represent the usual $not$ and $and$ operators for multi-valued logics.
The $b10m$ constraint is the multiplication of digits from 0 to 9,
i.e., $b10m(X,Y,C,Z)$ stands for the constraint $X*Y=Z+10*C$
defined over the intervals $[0..9]$.
The ``$-$'' symbol means that we were unable to generate the rules in less than
24 hours. 

The constraints for multi-valued logics are presented in a way that
shows the impact of the domain size and of the cardinality of the
constraint in case of the same arity and a similar structure. 

In spite of its exponential running time, the {\sc Equality Rules
Generation} algorithm is still usable. On the other hand, the {\sc
Membership Rules Generation} algorithm is much more costly for larger
problems. Sometimes it also generates too many rules for medium
size problems (such as $and_9$) and thus becomes unusable.

In general, it is difficult to decide which notion of local
consistency should be used to solve a given CSP. In particular
\cite{sabin-contradicting} showed that in the case of CSP's consisting
of binary constraints maintaining full arc consistency during the
backtracking search can be often more efficient than a more limited of
constraint propagation embodied in the so-called forward checking.
However, empirical results for CSP's involving non-binary are missing
and it is quite conceivable that for such CSP's imposing full arc
consistency during the backtracking search can be too costly.  For
these CSP's a weaker form of constraint propagation, such as rule
consistency, could be an alternative.

\end{document}